\documentclass{article}

    \PassOptionsToPackage{numbers}{natbib}

 \usepackage[main, final]{neurips_2026}

\usepackage[utf8]{inputenc} 
\usepackage[T1]{fontenc}    
\usepackage{hyperref}       
\usepackage{url}            
\usepackage{booktabs}       
\usepackage{amsfonts}       
\usepackage{nicefrac}       
\usepackage{microtype}      
\usepackage{xcolor}         

\newcommand{\name}{{\sc Drape }}
\newcommand{\mame}{{\sc Drape}}

\usepackage{algorithm}
\usepackage{algpseudocode}
\usepackage{subcaption}
\usepackage{multirow}
\usepackage{amsmath,amssymb}
\usepackage{mathtools}
\usepackage{xfrac}
\usepackage[english]{babel}
\usepackage{hyphenat}
\usepackage{marvosym}
\usepackage{tabularx} 
\usepackage{xcolor} 
\usepackage{colortbl}
\usepackage{soul}
\usepackage{csquotes}
\usepackage{listings}

\usepackage{seqsplit}
\usepackage{arydshln}
\usepackage{nicematrix}

\usepackage{xspace}
\newcommand{\ie}{\textit{i.e.}\@\xspace}

\usepackage{wrapfig}
\definecolor{Gray}{gray}{0.85}
\definecolor{Redo}{rgb}{0.95,0.69,0.51}
\definecolor{LightCyan}{rgb}{0.88,1,1}
\usepackage{afterpage}
\usepackage{pifont}
\usepackage{csquotes}
\usepackage{lipsum}

\title{Dynamic Cross-Modal Prompt Generation for\\ Multimodal Continual Instruction Tuning}

\author{Tao Hu$^{1,2}$  \ \ \
Da-Wei Zhou$^{1,2}\textsuperscript{(\Letter)}$ \\
$^{1} $ School of Artificial Intelligence, Nanjing University \\
$^{2} $ State Key Laboratory for Novel Software Technology, Nanjing University \\
\texttt{\small \{hut, zhoudw\}@lamda.nju.edu.cn}\\
}
\begin{document}
\maketitle

\begin{abstract}

Multimodal Large Language Models (MLLMs) achieve strong performance through instruction tuning, yet real-world deployment often requires continual capability expansion across sequential tasks. In such scenarios, Multimodal Continual Instruction Tuning (MCIT) aims to acquire new capabilities while limiting catastrophic forgetting. Existing methods mainly follow a module-composition paradigm: they maintain task-level prompts or LoRA experts and dynamically route or aggregate a subset of them at inference. However, samples within the same task can still differ substantially in visual scenes, question intents, and reasoning demands. This motivates instance-level adaptation to individual query-image pairs rather than only selecting or combining task-level modules. To this end, we propose \name (Dynamic Cross-Modal Prompt Generation), a prompt-learning framework that synthesizes continuous instance-specific soft prompts for MCIT. Instead of selecting prompts from a fixed pool, \name derives prompt queries from the textual instruction and cross-attends to visual patch features, producing prompts conditioned on query-image pairs that are prepended to the frozen LLM. To mitigate forgetting during sequential updates, \name applies null-space gradient projection to the shared projector and uses CLIP-based prototype routing for task-label-free generator selection at inference. Extensive experiments on MCIT benchmarks show that \name achieves state-of-the-art performance among representative prompt-based and LoRA-based continual-learning baselines. Code is available at: \url{https://github.com/LAMDA-CL/NeurIPS2026-Drape}.

\end{abstract}
    
\section{Introduction}
\label{sec:intro}

Recent Multimodal Large Language Models (MLLMs)~\cite{liu2024visual, dai2023instructblip} achieve strong performance through large-scale multimodal instruction tuning~\cite{zhang2026instruction,tong2025metamorph}, enabling them to handle diverse vision-language tasks~\cite{lu2022learn, goyal2017making}. Yet practical deployment rarely stops at a fixed task mixture: new domains, instruction styles, and capabilities arrive over time, and the model must absorb them without sacrificing what it has already learned. This setting gives rise to Multimodal Continual Instruction Tuning (MCIT)~\cite{chen2024coin}, where the central challenge is to extend multimodal capability under sequential updates while limiting catastrophic forgetting~\cite{french1999catastrophic, zhai2023investigating}.

Existing MCIT methods mainly follow a module-composition paradigm. They freeze the pre-trained LLM and vision backbone, maintain lightweight task-specific modules such as prompts or LoRA experts, and dynamically route or combine these modules at inference~\cite{cao2024continual, chen2024coin, yu2025progressive, guo2025hide,ge2025dynamic}. Although such routing or retrieval mechanisms make the effective configuration input-dependent, the adaptation remains restricted to selecting or recombining discrete modules learned from previous tasks. In practice, however, samples within the same MCIT task can still differ substantially in visual scenes, question intents, reasoning demands, and instruction--image interactions. These differences require the model to dynamically adapt to individual query--image pairs rather than only selecting or recombining task-level components. This motivates an alternative approach that synthesizes continuous prompts directly from the current multimodal input.

\begin{figure*}[t]
	\centering
	\begin{subfigure}{0.49\linewidth}
         \centering
		\includegraphics[width=0.8\columnwidth]{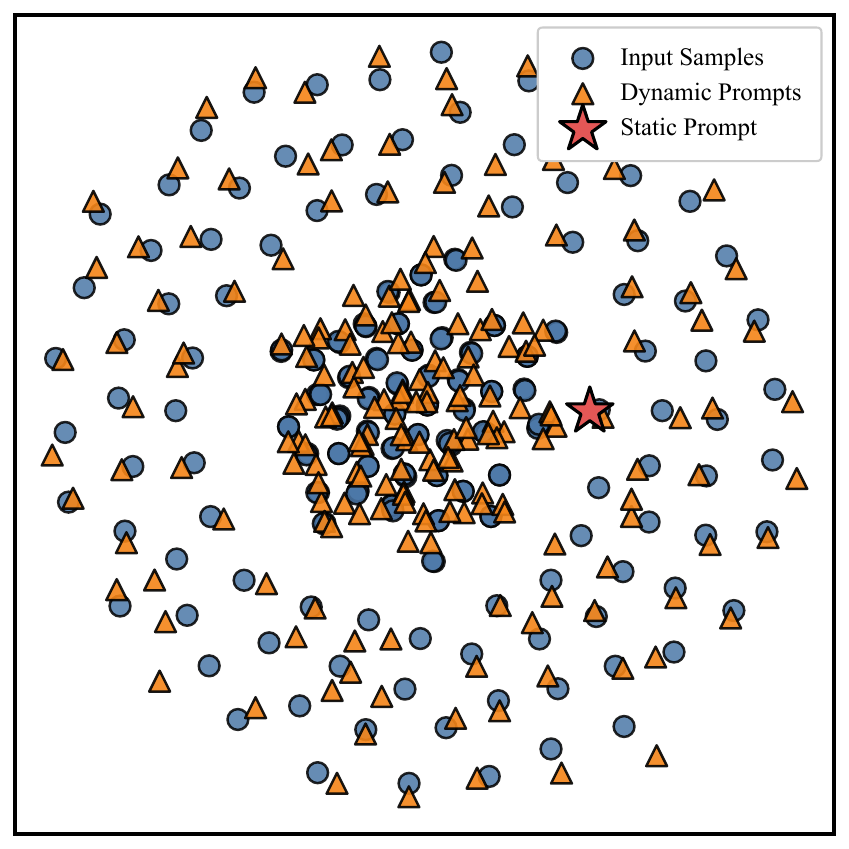}
	\end{subfigure}
	\begin{subfigure}{0.49\linewidth}
         \centering
		\includegraphics[width=0.95\columnwidth]{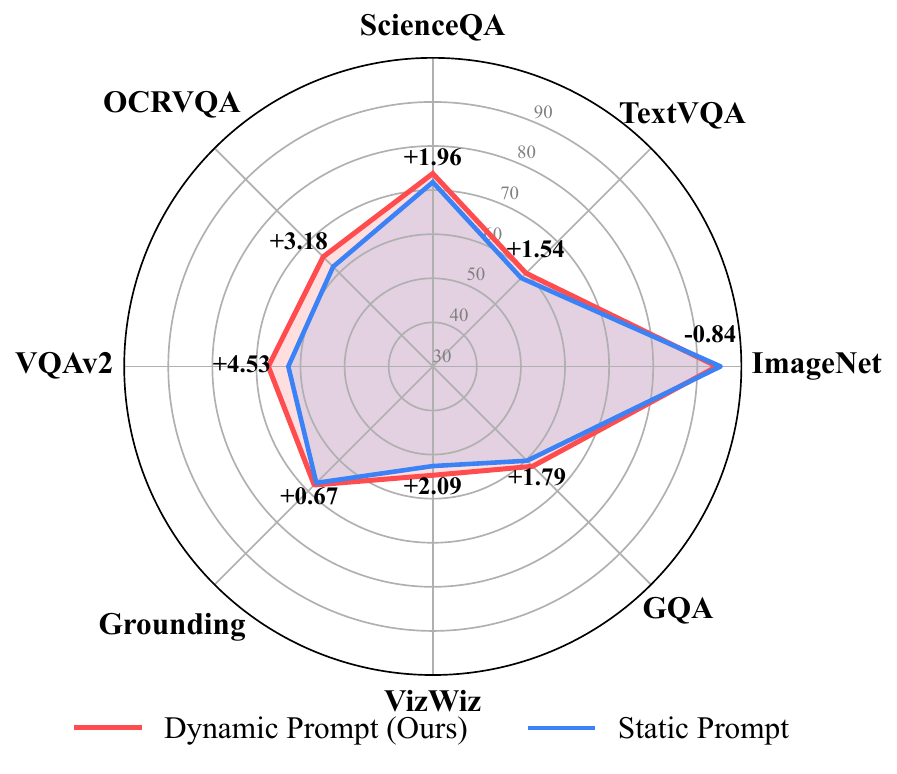}
	\end{subfigure}
	\caption{Why instance-specific prompts are needed. \textbf{Left:} t-SNE visualization of 200 samples from the VQAv2 test set. For each sample, we average-pool the input embeddings and the corresponding instance-specific soft prompt to obtain visualization features. For the static baseline, a single shared task prompt is average-pooled and plotted for comparison. The instance-specific soft prompts follow the sample distribution more closely, suggesting that they better reflect instance-level variation than static task prompts. \textbf{Right:} task-wise performance comparison between instance-specific and static task prompts on CoIN~\cite{chen2024coin}. Instance-specific prompts achieve stronger overall performance, suggesting the benefit of modeling within-task variation rather than using one fixed task-level prompt.}
	\label{fig:motivation}

\end{figure*}
This instance-level variation is further illustrated in Figure~\ref{fig:motivation}. In the left panel, the instance-specific soft prompts generated for each sample by our cross-modal prompt generator follow the sample distribution more closely than static task prompts, suggesting that useful adaptation should track instance-level structure rather than collapse all samples in a task into the same prompt. In the right panel, instance-specific prompts achieve stronger overall performance than static task prompts on CoIN~\cite{chen2024coin}, further suggesting that within-task multimodal diversity cannot be fully captured by one fixed task-level representation. These observations motivate a continual-learning framework that remains lightweight like prior parameter-efficient methods while allowing the adaptation itself to vary across samples.\looseness -1

To address this issue, we propose \name (Dynamic Cross-Modal Prompt Generation), a prompt-learning framework that generates continuous instance-specific soft prompts for MCIT. Although instance-dependent prompting has been explored in prior language-only or vision-only settings~\cite{xiao2025visual,lerevisit,wu2022idpg,liu2025all}, these methods do not directly model query--image conditioned adaptation under sequential multimodal instruction tuning. In contrast, \name synthesizes prompts from both textual instructions and visual patch features: a lightweight cross-modal prompt generator summarizes instruction tokens into prompt queries, cross-attends to visual patch features, and produces prompts that are prepended to the frozen LLM. Rather than selecting or recombining discrete task-level components, the model synthesizes prompts that adapt to the fine-grained multimodal structure of each sample.

To reduce forgetting during sequential updates, we combine this prompt generator with two complementary mechanisms. First, we apply null-space gradient projection to the shared projector so that updates are biased toward directions that minimally interfere with previously observed feature subspaces. Second, we use a CLIP-based prototype router to select the most relevant task-specific generator without task labels at inference. Together, these components form an input-conditioned cross-modal prompt-generation framework for rehearsal-free MCIT, improving adaptation to diverse query--image pairs while limiting catastrophic forgetting throughout training over sequential tasks.

\section{Related Work}
\label{sec:related_work}
\noindent\textbf{Multimodal Large Language Models.}
Multimodal Large Language Models (MLLMs) extend large language models to vision-language tasks by connecting visual encoders with language models through cross-modal alignment modules~\cite{radford2021learning,liu2024visual,dai2023instructblip}. Typically, visual features are projected into the language embedding space and combined with textual instructions, enabling the model to generate responses conditioned on both modalities~\cite{liu2024visual,dai2023instructblip}. With large-scale multimodal instruction tuning~\cite{zhang2026instruction,wei2021finetuned,longpre2023flan}, MLLMs have achieved strong performance on visual reasoning~\cite{johnson2017clevr,zerroug2022benchmark}, visual question answering~\cite{goyal2017making}, instruction following~\cite{zhou2023instruction}, and multimodal generation~\cite{dai2023instructblip}. Despite these advances, most existing MLLMs are trained on a fixed mixture of tasks and datasets~\cite{zhang2024mm,chen2024coin}. As a result, adapting them to newly emerging domains, instruction styles, or capabilities often requires additional fine-tuning, which may interfere with previously learned multimodal knowledge~\cite{shi2021overcoming,zhai2023investigating}. This static-training assumption motivates studying how MLLMs can be adapted after deployment.

\noindent\textbf{Continual Instruction Tuning.}
Continual instruction tuning extends standard instruction tuning by enabling foundation models to learn a sequence of tasks while remaining aligned with human instructions~\cite{guo2025comprehensive}. As MLLMs are increasingly deployed in open-world settings, mitigating catastrophic forgetting~\cite{zhai2023investigating} in Multimodal Continual Instruction Tuning (MCIT) has become an important problem~\cite{wu2024continual,zhang2024mm}. Existing methods generally follow several directions. Early approaches often adapt continual learning techniques from natural language processing~\cite{wang2024inscl,wang2024rehearsal,wang2023orthogonal}, using regularization or replay to preserve past knowledge. Recent parameter-efficient methods, such as MoELoRA~\cite{chen2024coin}, ProgLoRA~\cite{yu2025progressive}, and HiDe-LLaVA~\cite{guo2025hide}, keep the MLLM backbone frozen and update only lightweight task-specific modules. More recent studies further improve expert routing and knowledge reuse. SAME~\cite{xie2026same} stabilizes expert routing and updates to mitigate router and expert drift, while CRAM~\cite{tang2026cram} combines centroid-guided routing with adaptive expert allocation to balance knowledge reuse and parameter efficiency. ProtoAda~\cite{shi2026protoada} introduces format-aware task prototypes and geometric consolidation to reduce interference across heterogeneous response formats. ReCAP~\cite{hu2026recap} further leverages external knowledge retrieval to construct instance-specific capability paths from reusable modules. These approaches primarily address continual capability preservation and reuse through expert routing, parameter adaptation, or module composition. In parallel, instance-aware prompting has also been studied in several related settings, including instance-dependent prompt generation for NLP~\cite{wu2022idpg}, visual instance-aware or input-adaptive prompt tuning~\cite{xiao2025visual,lerevisit}, compact capsule prompt representations~\cite{liu2025all}, and image-conditioned prompt generation for continual learning~\cite{jung2023generating}.

\section{Preliminaries}
\label{sec:prelim}
\subsection{Multimodal Continual Instruction Tuning}
We consider a typical MLLM~\cite{liu2024improved} that processes an image $\mathbf{v}$ and a textual instruction $\mathbf{q}$ into a concatenated multimodal sequence $\mathbf{z}=[\mathbf{w};\mathbf{u}]\in\mathbb{R}^{(m+s)\times d}$. Here, $\mathbf{w}=\pi(\phi(\mathbf{v}))$ denotes the projected visual features from a vision encoder $\phi(\cdot)$ and a projector $\pi(\cdot)$, while $\mathbf{u}=\psi(\mathbf{q})$ represents the text embeddings. In the continual instruction tuning setting, the model encounters a sequence of tasks $\{\mathcal{D}_1, \mathcal{D}_2, \dots, \mathcal{D}_T\}$ arriving sequentially, where each task $\mathcal{D}_t = \{(\mathbf{v}_i, \mathbf{q}_i, \mathbf{y}_i)\}_{i=1}^{N_t}$ consists of image--instruction--answer triplets, and $\mathbf{y}_i$ denotes the target answer token sequence. Under the strict rehearsal-free constraint (\ie, past data is inaccessible), the goal is to optimize a set of tunable parameters $\Theta$ to generalize across all tasks seen so far:
\begin{equation}
\label{eq:cl_obj}
    \Theta_t^{*}
    = \arg\min_{\Theta}\;
    \mathbb{E}_{(\mathbf{v},\mathbf{q},\mathbf{y})\,\sim\, \mathcal{D}_{\leq t}}
    \!\left[
    -\sum_{j=1}^{L}
    \log\, p_\Theta\!\left(y_j \mid \mathbf{z},\, \mathbf{y}_{<j}\right)
    \right],
\end{equation}
where $\mathcal{D}_{\leq t}$ denotes the joint distribution over all observed tasks, $L$ is the length of the target answer sequence $\mathbf{y}$, and $y_j$ denotes its $j$-th token. In practice, since training is performed exclusively on $\mathcal{D}_t$, the learning procedure must inherently mitigate catastrophic forgetting.

\subsection{Baselines in MCIT}
To overcome forgetting while maintaining the pre-trained multimodal knowledge, existing rehearsal-free MCIT methods typically freeze the LLM backbone and rely on two dominant paradigms: Prompt Tuning and Mixture-of-Experts (MoE) with LoRA.

\noindent\textbf{Prompt Tuning:} Prompt-based methods~\cite{smith2023coda,wang2022learning,zeng2025modalprompt} adapt the frozen model by prepending learnable continuous vectors (soft prompts) to the input sequence. To handle a continuous sequence of tasks, they typically maintain a dynamic prompt pool $\mathcal{P}$. For a given input, specific prompts $\mathbf{P}\in\mathbb{R}^{L_p\times d}$ are retrieved or matched based on instance features. The augmented input becomes $\mathbf{z}'=[\mathbf{P};\mathbf{z}]$, and the retrieved prompts, or task-specific prompts when task identities are used, are updated during training.\looseness -1

\noindent\textbf{MoE with LoRA Experts:} An alternative paradigm~\cite{chen2024coin} focuses on parameter-efficient fine-tuning via Low-Rank Adaptation (LoRA). To prevent forgetting, these methods allocate task-specific LoRA experts and freeze them after learning each task. During inference, a routing mechanism dynamically blends these historical experts. For a given input, the dynamically assembled parameter $\Theta_{add}$ is computed as:
\begin{equation}\label{eq:assembly}
\Theta_{add} = \sum_{k=1}^{t} \omega_k \Theta_k,
\end{equation}
where $\Theta_k$ represents the LoRA parameters optimized for task $k$, and $\omega_k$ is the routing weight predicted by a gating network based on current input features.

\noindent\textbf{Discussion.}
Both paradigms provide practical solutions for MCIT, but they also expose trade-offs.
For prompt tuning, the reliance on a discrete prompt pool $\mathcal{P}$ can limit how flexibly the adaptation responds to within-task variation.
For MoE with LoRA experts, the model gains stronger task-level specialization, but its behavior is still governed by routing and recombination among stored task-specific components.
These approaches can be effective, yet they may be less suitable when subtle multimodal differences call for more input-specific adaptation.
This motivates exploring whether MCIT can benefit from dynamically synthesizing prompts for each sample while still preserving historical knowledge through lightweight constraints.
\begin{figure*}[t]
	\centering
      {\includegraphics[width=1\columnwidth]{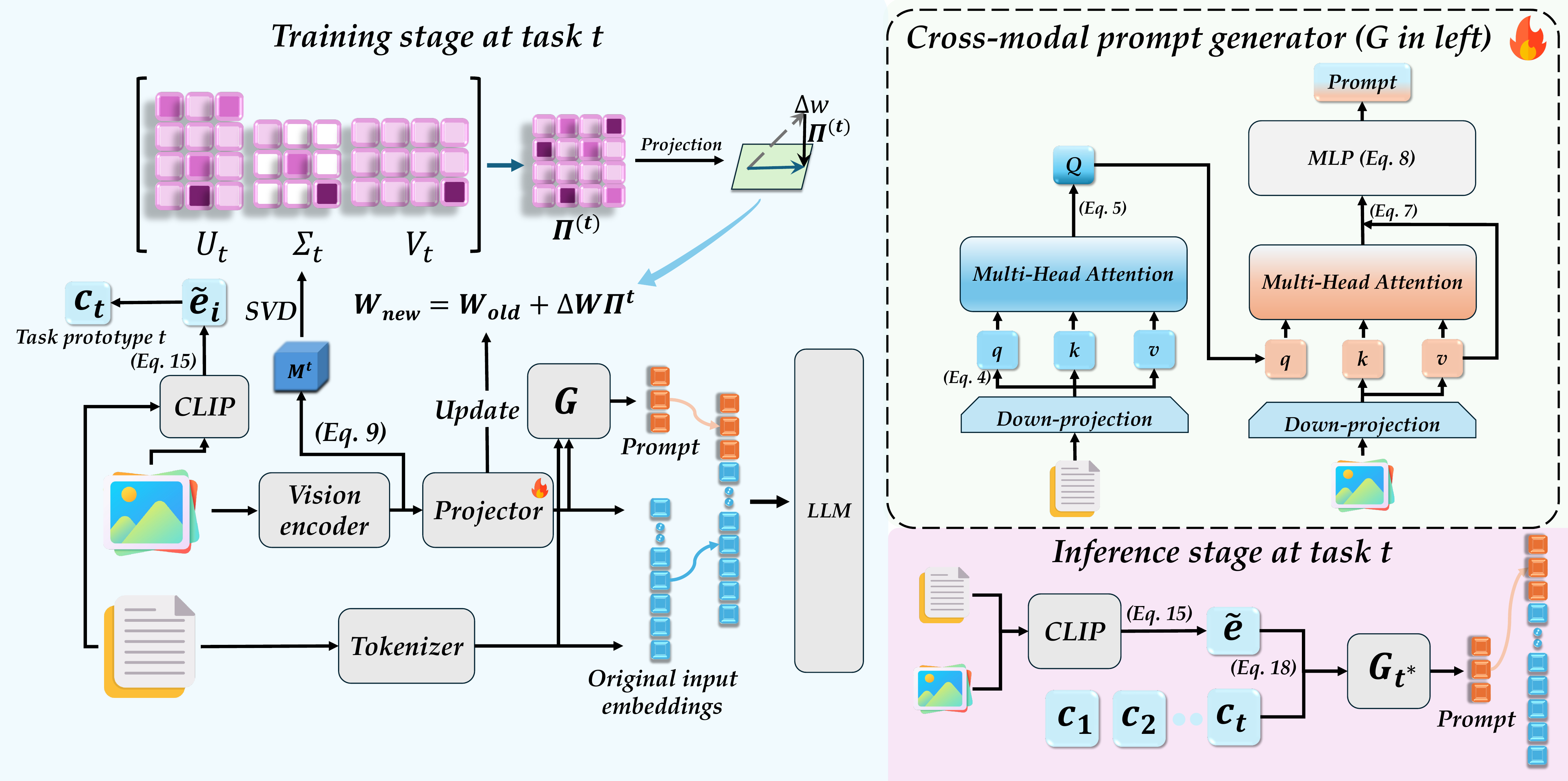}}
	\caption{Illustration of \mame. {\bf Left}: Training on task $t$. A task-specific generator synthesizes soft prompts, while the shared visual projector is regularized by projection onto the complement of the retained principal subspace. Feature statistics $\mathbf{M}^{(t)}$ are decomposed via SVD to obtain a projection matrix $\boldsymbol{\Pi}^{(t)}$ for the next task. After training, a task prototype $\mathbf{c}_t$ is registered in a frozen CLIP embedding space. {\bf Top-Right}: Cross-modal prompt generator. It summarizes instruction tokens into prompt queries and cross-attends to visual features to synthesize instance-specific soft prompts. {\bf Bottom-Right}: Inference without task labels. The router compares the current input against stored prototypes and selects the corresponding generator $G_{t^*}$.}
	\label{figure:teaser}

\end{figure*}
\section{{\scshape{Drape}}: Dynamic Cross-Modal Prompt Generation}
\label{sec:method}

We propose \mame, a prompt-learning framework for MCIT that moves from static component reuse toward input-conditioned prompt synthesis. As illustrated in Figure~\ref{figure:teaser}, \name contains three key components. First, a cross-modal prompt generator produces instance-specific soft prompts conditioned on both the visual and textual features of each input. Second, null-space gradient projection constrains updates to the shared visual projector so as to reduce interference with previously observed feature subspaces. Third, a CLIP-guided prototype router selects the most relevant task-specific generator at inference. In the main experiments, the router stores one learned prototype per task, built from fused CLIP text-image features.

\subsection{Cross-Modal Prompt Generation}
\label{subsec:generator}

The core of \name is a cross-modal prompt generator $G_t$ that, given an input's projected visual features $\mathbf{w}\in\mathbb{R}^{m\times d}$ and textual instruction embeddings $\mathbf{u}\in\mathbb{R}^{s\times d}$, synthesizes $L_p$ instance-specific soft prompt vectors $\mathbf{P}_i = G_t(\mathbf{w}_i, \mathbf{u}_i)\in\mathbb{R}^{L_p\times d}$. The subscript $t$ indicates that a separate task-specific generator is maintained for each task. During training, the generator masks out image placeholders, padding tokens, and answer positions so that prompt synthesis depends on the non-target instruction context rather than the supervised response. In our default configuration, the generation process uses two stages: instruction-aware query initialization followed by vision-guided prompt synthesis.

\noindent\textbf{Instruction-Aware Query Initialization.}
The textual instruction encodes the task intent (e.g., ``describe this image'' vs.\ ``what is the OCR text'') and should guide how visual information is extracted. We first project the masked instruction embeddings into a latent space via a linear mapping $f_u:\mathbb{R}^{d}\to\mathbb{R}^{H}$:
\begin{equation}
    \mathbf{h}^{u} = f_u(\mathbf{u}) \in \mathbb{R}^{s \times H},
\end{equation}
where $H$ is the generator's hidden dimension. In the default `segment' pooling mode, the instruction sequence is partitioned into $L_p$ roughly equal segments $\{\mathcal{S}_1, \dots, \mathcal{S}_{L_p}\}$, and a masked average pooling is computed within each segment:
\begin{equation}
    \bar{\mathbf{h}}_p = \frac{\sum_{j\in\mathcal{S}_p} \mathbf{h}^{u}_j \cdot \mathbb{1}[j\text{ is valid}]}{\max\bigl(|\{j\in\mathcal{S}_p : j\text{ is valid}\}|,\, 1\bigr)}, \quad p=1,\dots,L_p,
\end{equation}
yielding an initial pooled query matrix $\bar{\mathbf{H}} \in \mathbb{R}^{L_p \times H}$. These pooled summaries are then used as queries in a multi-head attention layer over the full instruction features:
\begin{equation}
    \mathbf{Q} = \mathrm{LN}\!\left(\mathrm{MHA}(\bar{\mathbf{H}},\, \mathbf{h}^{u},\, \mathbf{h}^{u})\right) \in \mathbb{R}^{L_p \times H},
\end{equation}
where $\mathrm{MHA}$ denotes multi-head attention with a key padding mask that ignores invalid positions, and $\mathrm{LN}$ denotes layer normalization. The resulting $\mathbf{Q}$ contains $L_p$ instruction-aware query vectors that are passed to the visual cross-attention stage.

\noindent\textbf{Vision-Guided Prompt Synthesis.}
With the instruction-derived queries in hand, we extract task-relevant visual information through cross-modal attention. The projected visual features $\mathbf{w}$ are first mapped to the latent space:
\begin{equation}
    \mathbf{h}^{v} = f_v(\mathbf{w}) \in \mathbb{R}^{m \times H},
\end{equation}
where $f_v:\mathbb{R}^{d}\to\mathbb{R}^{H}$ is a linear projection. The instruction queries $\mathbf{Q}$ then attend to the visual features via cross-modal attention:
\begin{equation}\label{eq:cross_attn}
    \mathbf{R} = \mathrm{LN}\!\left(\mathbf{Q} + \mathrm{MHA}(\mathbf{Q},\, \mathbf{h}^{v},\, \mathbf{h}^{v})\right) \in \mathbb{R}^{L_p \times H}.
\end{equation}
A dropout layer is applied to $\mathbf{R}$ for regularization. Finally, a two-layer MLP projects the fused representations back to the LLM's embedding space, producing the instance-specific soft prompts:
\begin{equation}
    \mathbf{P}_i = f_{\mathrm{head}}(\mathbf{R}) \in \mathbb{R}^{L_p \times d},
\end{equation}
where $f_{\mathrm{head}}$ maps $\mathbb{R}^{H}\to\mathbb{R}^{2H}\to\mathbb{R}^{d}$ with a GELU activation in between. The instance-specific soft prompts are prepended to the multimodal input sequence, yielding $\mathbf{z}' = [\mathbf{P}_i;\, \mathbf{z}]$, and the LLM generates the response via $p_\theta(\mathbf{y}\mid\mathbf{z}')$.
\\\textbf{Discussion.}
This design choice reflects an asymmetry between modalities: the instruction indicates \emph{which} information is needed, whereas the image provides the content that must be adapted to. By initializing prompt queries from the instruction and refining them through visual cross-attention, \name performs input-conditioned prompt adaptation instead of selecting from a fixed pool of historical components. The implementation also supports mean-pooling, static-prompt, and learnable-query variants for additional analysis beyond the default setting.

\subsection{Null-Space Gradient Projection for Projector Preservation}
\label{subsec:nullspace}

While each task maintains its own generator, the visual projector $\pi(\cdot)$ is shared across all tasks and continuously updated during sequential training. Updates from later tasks may therefore alter the visual-to-language mappings used by earlier tasks and contribute to catastrophic forgetting. To mitigate this effect, we constrain the gradient updates of $\pi$ using an approximate null-space projection derived from the empirical feature subspaces observed in previous tasks.

Concretely, during training on task $\mathcal{D}_t$, forward hooks attached to each linear layer of $\pi$ collect the input features $\tilde{\mathbf{v}}$ passing through that layer. The statistics and projections below are maintained per linear layer, with layer indices omitted for clarity. After training completes, these features are merged with the statistics accumulated from previous tasks to form a cumulative second-moment matrix:
\begin{equation}
    \mathbf{M}^{(t)} = \frac{N^{(t-1)}\,\mathbf{M}^{(t-1)} + \sum_{i}\tilde{\mathbf{v}}_i \tilde{\mathbf{v}}_i^\top}{N^{(t)}}\,,\quad N^{(t)} = N^{(t-1)} + N_t\,,
\end{equation}
where $N_t$ is the number of feature vectors collected from $\mathcal{D}_t$ for the corresponding layer and $\mathbf{M}^{(0)} = \mathbf{0}$. We then decompose $\mathbf{M}^{(t)}$ via singular value decomposition:
\begin{equation}
    \mathbf{M}^{(t)} = U\,\Sigma\,V^\top, \quad \Sigma = \mathrm{diag}(\sigma_1 \geq \cdots \geq \sigma_{d_v}).
\end{equation}
The leading singular directions provide an empirical estimate of the dominant feature subspace used by the tasks observed so far. We identify the effective rank $r$ as the smallest index satisfying:
\begin{equation}
    \frac{\sum_{k=1}^{r} \sigma_k}{\sum_{k=1}^{d_v} \sigma_k} \geq \epsilon\,,
\end{equation}
where $\epsilon$ is the energy-retention threshold. To preserve at least one update direction, we enforce $r < d_v$. The directions outside the retained subspace form the complementary basis
$V_\perp = V_{[:,\,r+1:d_v]}$, and we construct the projection matrix:
\begin{equation}
    \boldsymbol{\Pi}^{(t)} = V_\perp\, V_\perp^\top \in \mathbb{R}^{d_v \times d_v}.
\end{equation}
During training on task $\mathcal{D}_{t+1}$, $\boldsymbol{\Pi}^{(t)}$ is applied to the gradients of each weight matrix in $\pi$ via registered gradient hooks. For a weight matrix $W \in \mathbb{R}^{d \times d_v}$, the projected gradient is:
\begin{equation}\label{eq:grad_proj}
    \nabla W^\prime = \nabla W \boldsymbol{\Pi}^{(t)} = \nabla W V_\perp V_\perp^\top.
\end{equation}
This operation restricts the projector update to the complementary subspace of the dominant feature directions accumulated from previous tasks. To see why this helps preserve earlier-task mappings, consider an earlier-task input feature $\tilde{\mathbf{v}}^{\mathrm{old}}$ to the same linear layer. If this feature is well captured by the retained principal subspace, then its component in the complementary subspace is small, i.e., $V_\perp^\top \tilde{\mathbf{v}}^{\mathrm{old}} \approx \mathbf{0}$. For an idealized gradient descent step with learning rate $\eta$, let $\Delta \mathbf{o}^{\mathrm{old}}$ denote the first-order change in the layer output for this old feature. Using the projected gradient in Eq.~\eqref{eq:grad_proj}, we have:
\begin{equation}
    \Delta \mathbf{o}^{\mathrm{old}}
    =
    -\eta \nabla W^\prime \tilde{\mathbf{v}}^{\mathrm{old}}
    =
    -\eta \nabla W V_\perp
    \left(
    V_\perp^\top \tilde{\mathbf{v}}^{\mathrm{old}}
    \right)
    \approx \mathbf{0}.
\end{equation}
Thus, when previous-task features mainly lie in the retained principal subspace, the projected update has limited effect on their projected representations. In this way, the projection helps reduce forgetting in the shared projector while still allowing updates along complementary directions for new tasks.

\noindent\textbf{Discussion.}
The energy threshold $\epsilon$ determines how many dominant feature directions from previous tasks are retained. A larger $\epsilon$ preserves more historical feature directions, making subsequent projector updates less likely to overwrite visual-to-language mappings used by earlier tasks. In contrast to replay-based methods, this strategy stores only compact second-moment statistics for the shared projector. Thus, it provides a lightweight complement to the task-specific generators for reducing forgetting without retaining past training samples.

\subsection{CLIP-Guided Prototype Routing}
\label{subsec:router}

Since \name maintains a separate generator $G_t$ for each task, a routing mechanism is required at inference to select the appropriate generator without task labels. We use frozen CLIP embeddings as a shared semantic space for routing. In particular, the visual features are directly reused from the CLIP-based vision tower of LLaVA, so the router does not require loading an additional CLIP encoder, introducing only marginal extra cost for text-side encoding.

\noindent\textbf{Task Prototype Computation.}
After learning each task, we register one compact prototype in the CLIP space. The routing feature is constructed by concatenating the normalized CLIP text embedding of the instruction and the normalized CLIP image embedding:
\begin{equation}
    \tilde{\mathbf{e}}_i = \mathrm{norm}\!\left([\xi(\mathbf{q}_i);\, \gamma(\mathbf{v}_i)]\right).
\end{equation}
During training, fused embeddings from the current task are cached temporarily and used to register the task prototype once training on $\mathcal{D}_t$ is finished. Let $\{\tilde{\mathbf{e}}_i\}_{i=1}^{N_t}$ denote the cached embeddings of task $t$. We initialize the prototype $\mathbf{c}_t$ by the normalized mean of these embeddings. For $t>1$, we further refine $\mathbf{c}_t$ with a classification objective that pulls current-task embeddings closer to $\mathbf{c}_t$ than to previously stored prototypes. For a minibatch $\{\tilde{\mathbf{e}}_i\}_{i=1}^{B}$ and earlier prototypes $\{\mathbf{c}_s\}_{s=1}^{t-1}$, the loss is:
\begin{equation}
    \mathcal{L}_{\mathrm{cls}} = -\frac{1}{B}\sum_{i=1}^{B}\log 
    \frac{\exp\!\left(\cos(\tilde{\mathbf{e}}_i,\mathbf{c}_t) / \tau\right)}
    {\exp\!\left(\cos(\tilde{\mathbf{e}}_i,\mathbf{c}_t) / \tau\right) + 
    \sum_{s=1}^{t-1} \exp\!\left(\cos(\tilde{\mathbf{e}}_i,\mathbf{c}_s) / \tau\right)},
    \label{eq:prototype_loss}
\end{equation}
where $\tau$ is the temperature parameter and $\cos(\cdot,\cdot)$ denotes cosine similarity. For the first task, no previous prototype is available, so Eq.~\ref{eq:prototype_loss} is not applied and $\mathbf{c}_1$ is simply registered as the normalized mean prototype. After registration, the cached instance-level embeddings are discarded. As a result, only one compact prototype is retained for each learned task.

\noindent\textbf{Inference Routing.}
Given a test sample, we compute its routing feature $\tilde{\mathbf{e}}$ in the same CLIP space and score each task by cosine similarity:
\begin{equation}
    s_t = \cos(\tilde{\mathbf{e}},\mathbf{c}_t).
\end{equation}
The router selects
\begin{equation}
    t^* = \arg\max_{t \in \{1,\dots,T\}} s_t,
\end{equation}
and activates the corresponding generator $G_{t^*}$ to produce the prompt.

\noindent\textbf{Discussion.}
The router is lightweight: it stores only one prototype per task and does not rely on replay samples or historical feature banks. Moreover, by reusing the CLIP-based vision tower already available in LLaVA, it avoids introducing an additional CLIP model. Using fused text-image embeddings makes routing depend on both instruction semantics and visual content, while the prototype update keeps each task representation compact yet discriminative.

\subsection{Summary of \name}
\label{subsec:summary}

\name addresses multimodal continual instruction tuning through three complementary mechanisms: a cross-modal prompt generator for input-conditioned prompt synthesis, a projection-based constraint that reduces interference in the shared visual projector, and a CLIP-guided prototype router for practical task-specific generator selection without task labels.

\noindent\textbf{Training.}
When learning task $\mathcal{D}_t$, only the current generator $G_t$ and the shared visual projector $\pi$ are updated. The vision encoder $\phi$, the LLM backbone $\theta$, and all historical generators $\{G_1,\dots,G_{t-1}\}$ remain frozen. The training objective is the standard autoregressive language modeling loss optimized over the tunable parameters $\Theta_t = \{G_t, \pi\}$:
\begin{equation}\label{eq:overall_loss}
    \mathcal{L}_t = -\sum_{j=1}^{L}\log\,p_\theta\!\left(y_j \mid [\mathbf{P}_i;\,\mathbf{z}],\, \mathbf{y}_{<j}\right),
\end{equation}
where $\mathbf{P}_i = G_t(\mathbf{w}_i, \mathbf{u}_i)$ is the synthesized prompt. For $t>1$, the null-space projection in Eq.~\eqref{eq:grad_proj} is applied to the gradients of $\pi$ via backward hooks. After completing $\mathcal{D}_t$, three post-training steps are executed: (1)~the current generator $G_t$ is frozen; (2)~the cumulative second-moment matrix $\mathbf{M}^{(t)}$ is updated and the new projection matrix $\boldsymbol{\Pi}^{(t)}$ is computed for the next task; and (3)~the task prototype $\mathbf{c}_t$ is registered in the router from the current task's temporary fused CLIP embeddings, after which those per-instance embeddings are discarded.

\noindent\textbf{Inference.}
Given a test input $(\mathbf{v}, \mathbf{q})$ without task labels, the router computes the fused CLIP feature $\tilde{\mathbf{e}}$, selects $t^*$ by prototype similarity, and activates $G_{t^*}$ to synthesize $\mathbf{P}_i = G_{t^*}(\mathbf{w}_i, \mathbf{u}_i)$. The frozen LLM then generates the response from $[\mathbf{P}_i;\,\mathbf{z}]$.

\begin{table*}[t]
\centering
\small
\setlength{\tabcolsep}{2pt}
\renewcommand{\arraystretch}{1.08}
\setlength{\extrarowheight}{1pt}
\caption{Main results on the CoIN benchmark with LLaVA-v1.5-7B as the backbone (higher is better). The best and second-best values are marked in \textbf{bold} and \underline{underline}, respectively.}
\label{tab:main}
\resizebox{\linewidth}{!}{
\begin{tabular}{l|cccccccc|c}
\hline
\rowcolor{gray!20}
\textbf{Methods} & ScienceQA & TextVQA & ImageNet & GQA & VizWiz & Grounding & VQAv2 & OCR-VQA & Average \\
\hline
Finetune & 26.00 & 25.38 & 28.51 & 33.07 & 26.52 & 0.10 & 40.00 & 52.92 & 29.06 \\
\rowcolor{gray!10} CODA-Prompt~\cite{smith2023coda} & 58.15 & 50.16 & 24.04 & 54.33 & 48.94 & 17.83 & 55.86 & 54.42 & 45.46 \\
DualPrompt~\cite{wang2022dualprompt} & 56.40 & 47.12 & 34.96 & 42.03 & 44.14 & 12.01 & 54.43 & 53.36 & 43.05 \\
\rowcolor{gray!10} L2P~\cite{wang2022learning}& 54.42 & 46.04 & 30.36 & 57.09 & 42.19 & 9.38 & 50.45 & 54.03 & 42.99 \\
MoELoRA~\cite{chen2024coin} &47.34& 32.91 &38.73 &37.15 &42.48& 0.97 &42.77& 57.50&37.48 \\
\rowcolor{gray!10} Continual LLaVA~\cite{cao2024continual} & 58.67 & 49.99 & {57.66} & \textbf{62.53} & 42.32 & 16.25 & 64.33 & \textbf{74.91} & 53.33 \\
\rowcolor{gray!10}ModalPrompt~\cite{zeng2025modalprompt} & 68.42 & \underline{56.40} & 41.13 & {61.11} & {50.13} & \underline{36.69} & \textbf{66.90} & 59.68 & 55.06 \\
ProgLoRA~\cite{yu2025progressive} & \textbf{74.84} & 51.83 & \underline{83.90} & 49.93 & \underline{53.87} & 31.19 & 62.71 & {64.44} & \underline{59.09} \\
\hline
\rowcolor{LightCyan} \textbf{\name (Ours)}&\underline{70.67}& \textbf{59.61}& \textbf{94.16}& \underline{61.37}& \textbf{54.43}& \textbf{67.92}& \underline{66.53}& \underline{65.11}& \textbf{67.48}\\

\hline
\end{tabular}
}
\end{table*}

\begin{table*}[t]
\centering
\small
\setlength{\tabcolsep}{8pt}
\renewcommand{\arraystretch}{1.08}
\setlength{\extrarowheight}{1pt}
\caption{Main results on the UCIT benchmark with LLaVA-v1.5-7B as the backbone (higher is better). The best and second-best values are marked in \textbf{bold} and \underline{underline}, respectively.}
\label{tab:ucit-main}
\resizebox{\linewidth}{!}{
\begin{tabular}{l|cccccc|c}
\hline
\rowcolor{gray!20}
\textbf{Methods} & ImgNet-R & ArxivQA & VizWiz & IconQA & CLEVR & Flickr30k & Average \\
\hline
Zero-shot & 16.27 & 53.73 & 38.39 & 19.20 & 20.63 & 41.88 & - \\
\hline
LoRA-FT~\cite{hu2022lora} & 58.03 & 77.63 & 44.39 & \underline{67.40} & 61.77 & \textbf{58.22} & 61.24 \\
\rowcolor{gray!10} O-LoRA~\cite{wang2023orthogonal} & 77.50 & 78.07 & 44.50 & 63.13 & 64.73 & \underline{58.16} & 64.35 \\
MoELoRA~\cite{chen2024coin} & 70.07 & 77.70 & 44.69 & 50.03 & 54.03 & 57.34 & 58.98 \\
\rowcolor{gray!10} ModalPrompt~\cite{zeng2025modalprompt} & 51.07 & 87.27 & \underline{48.11} & 39.23 & 46.57 & 42.93 & 52.53 \\
CL-MoE~\cite{huai2025cl} & 66.33 & 77.00 & 44.78 & 51.87 & 53.53 & 57.42 & 58.49 \\
\rowcolor{gray!10} HiDe~\cite{guo2025hide} & \underline{84.03} & \underline{90.73} & 44.43 & 58.93 & 41.37 & 54.25 & 62.29 \\
SEFE~\cite{chen2025sefe} & 80.83 & 78.00 & 47.01 & \textbf{69.63} & \textbf{65.83} & 57.92 & \underline{66.54} \\
\hline
\rowcolor{LightCyan} \textbf{\name (Ours)} & \textbf{85.07} & \textbf{91.60} & \textbf{55.68} & 62.53 & \underline{65.77} & 55.82 & \textbf{69.41} \\
\hline
\end{tabular}
}
\end{table*}

\section{Experiments}
\label{sec:exp}

\subsection{Implementation Details}

\noindent{\bf Datasets.}
We conduct experiments on two MCIT benchmarks. The first is CoIN~\cite{chen2024coin}, which consists of eight sequential multimodal tasks: ScienceQA~\cite{lu2022learn}, TextVQA~\cite{singh2019towards}, ImageNet~\cite{deng2009imagenet}, GQA~\cite{hudson2019gqa}, VizWiz~\cite{gurari2018vizwiz}, Grounding (RefCOCO)~\cite{kazemzadeh2014referitgame, mao2016generation}, VQAv2~\cite{goyal2017making}, and OCR-VQA~\cite{mishra2019ocr}. The second is UCIT~\cite{guo2025hide}, which contains six sequential tasks: ImageNet-R~\cite{hendrycks2021many}, ArxivQA~\cite{li2024multimodal}, VizWiz-caption~\cite{gurari2018vizwiz}, IconQA~\cite{lu2021iconqa}, CLEVR-Math~\cite{lindstrom2022clevr}, and Flickr30k~\cite{plummer2015flickr30k}. Together, these two benchmarks let us evaluate our method in both a widely used MCIT setting and a cleaner setting with reduced data-overlap concerns.

{\bf Comparison Methods.}
We compare \name with classic prompt-based continual learning approaches, including CODA-Prompt~\cite{smith2023coda}, DualPrompt~\cite{wang2022dualprompt}, and L2P~\cite{wang2022learning}, as well as recent MCIT baselines, including MoELoRA~\cite{chen2024coin}, Continual LLaVA~\cite{cao2024continual}, ModalPrompt~\cite{zeng2025modalprompt}, and ProgLoRA~\cite{yu2025progressive}. For UCIT, we additionally include LoRA-FT~\cite{hu2022lora}, O-LoRA~\cite{wang2023orthogonal}, CL-MoE~\cite{huai2025cl}, HiDe~\cite{guo2025hide}, and SEFE~\cite{chen2025sefe}. We also report standard sequential fine-tuning as a forgetting-heavy lower bound.

{\bf Training Setup.}
We adopt LLaVA-v1.5-7B~\cite{liu2024visual} as the backbone MLLM, with CLIP-ViT-L/14-336~\cite{radford2021learning} as the vision encoder and a two-layer MLP with GELU activation as the visual projector. 
We conduct all experiments using Prism~\cite{tang2026prism}.
For the proposed generator, we set the hidden dimension to $H=512$ and the prompt length to $L_p=10$. The null-space threshold is $\epsilon=0.99$. Each task is trained for one epoch with per-device batch size $4$ and gradient accumulation steps $2$. We use cosine learning-rate decay with warmup ratio $0.03$, setting the peak learning rates to $2\times10^{-4}$ for the generator and $2\times10^{-5}$ for the projector. All experiments are conducted on 4 NVIDIA RTX 4090 GPUs with DeepSpeed ZeRO-2 and \texttt{bf16} precision.

{\bf Evaluation Metrics.}
Following~\cite{zhou2024class,zhou2024continual,chen2024coin}, we denote by $\mathcal{A}_{s,t}$ the performance on task $s$ after training up to task $t$, with $T$ total tasks. The primary metric is the final average accuracy, defined as $\Bar{\mathcal{A}}=\frac{1}{T}\sum_{s=1}^{T}\mathcal{A}_{s,T}$. We also report Backward Transfer (B) and Mean Accuracy (M), with their detailed formulations provided in  Appendix~\ref{app:evaluation-metrics}.

\subsection{Benchmark Comparison}
Table~\ref{tab:main} reports the benchmark comparison on CoIN, where \name achieves the highest average accuracy among all compared methods (\textbf{67.48}), outperforming ProgLoRA (59.09) by 8.39 points and obtaining the best results on four tasks: TextVQA, ImageNet, VizWiz, and Grounding. Table~\ref{tab:ucit-main} further shows that this advantage generalizes to UCIT, where \name again achieves the best average performance (\textbf{69.41}), surpassing the strongest baseline SEFE (66.54) by 2.87 points and ranking first on ImgNet-R, ArxivQA, and VizWiz. These results show that the gains of \name are robust across both the standard CoIN benchmark and the cleaner UCIT setting. Stage-wise forgetting analysis using Backward Transfer and Mean Accuracy is provided in  Appendix~\ref{app:continuous-performance}.

\begin{table}
\centering
\small
\setlength{\tabcolsep}{1.8pt}
\renewcommand{\arraystretch}{1.1}
\setlength{\extrarowheight}{1pt}
\caption{Ablation study on the CoIN benchmark. $\Delta$ indicates the average accuracy change relative to the full model. The best results are shown in \textbf{bold}.}
\label{tab:ablation}
\resizebox{\linewidth}{!}{
\begin{tabular}{l|cccccccc|cc}
\hline
\rowcolor{gray!20}
{Variants} & ScienceQA & TextVQA & ImageNet & GQA & VizWiz & Grounding & VQAv2 & OCR-VQA  & Average & {$\Delta$} \\
\hline
\rowcolor{LightCyan!50}{\name (Full)} & \textbf{70.67} & \textbf{59.61} & \textbf{94.16} & 61.37 & \textbf{54.43} & \textbf{67.92} & 66.53 & 65.11 & \textbf{67.48} & - \\
\hline
~~w/o Cross-Modal Attn.  & 69.39 & 59.44 & 94.14 & 61.11 & 47.77 & 67.87 & 66.55 & 64.22 & 66.31& -1.17 \\
~~w/o Null-Space Proj.& 69.94 & 55.78 & 83.62 & \textbf{61.48} & 49.83 & 65.43 & \textbf{67.43} & \textbf{65.38} & 64.86 & -2.62 \\
\hline
\end{tabular}}
\end{table}

\subsection{Further Analysis}

\noindent{\bf Ablation Study.} To evaluate the contribution of each core component in \mame, we conduct controlled ablations in Table~\ref{tab:ablation}. Removing the vision cross-attention stage (``w/o Cross-Modal Attn.'') reduces the average accuracy from 67.48 to 66.31, a drop of 1.17 points, with the clearest degradation appearing on visually demanding tasks such as VizWiz and OCR-VQA. Disabling null-space gradient projection (``w/o Null-Space Proj.'') causes a larger drop to 64.86, \ie, 2.62 points below the full model, with especially visible declines on ImageNet, TextVQA, VizWiz, and Grounding. This pattern is consistent with the intended role of the projector constraint: it does not eliminate forgetting entirely, but it helps reduce cross-task interference during sequential updates.

\begin{figure}[t]
    \centering
    \begin{subfigure}{0.64\linewidth}
        \includegraphics[width=\linewidth]{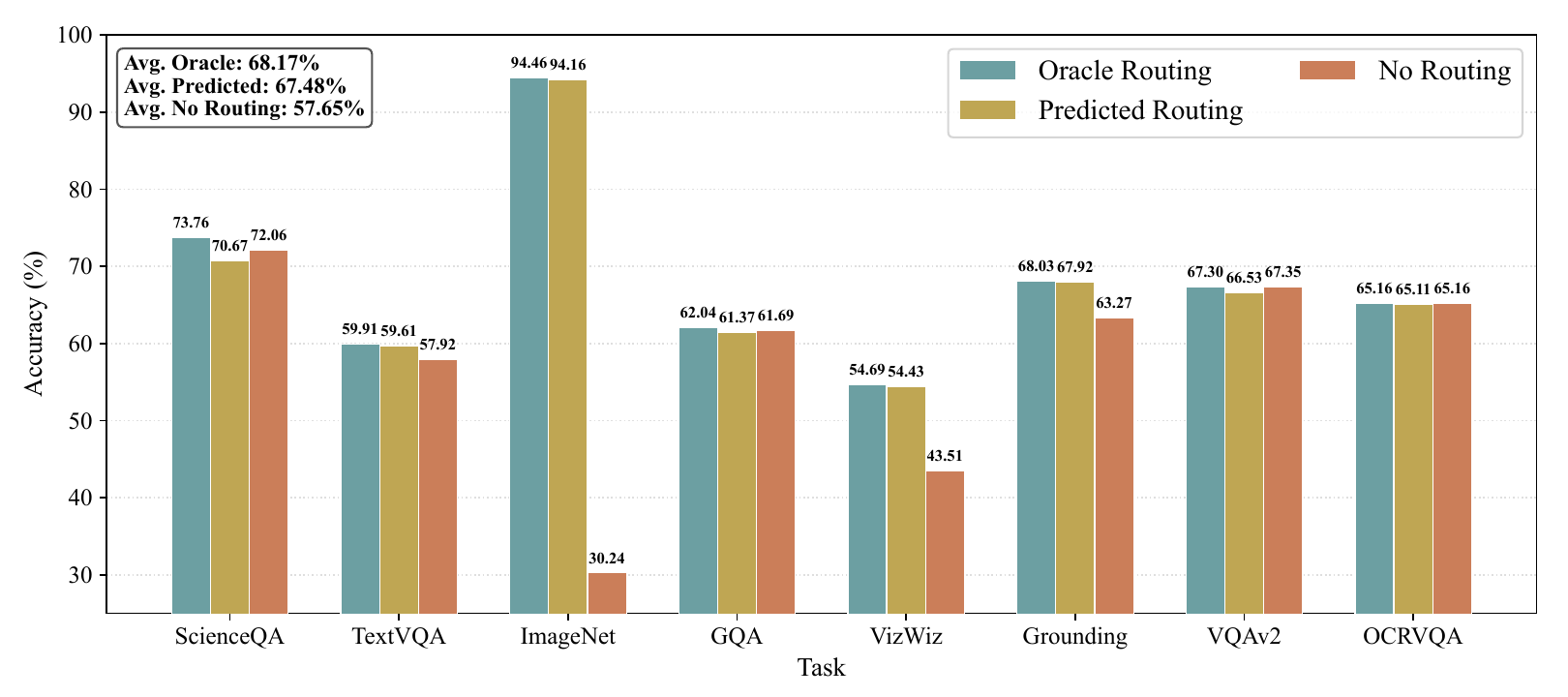}
    \end{subfigure}\hfill
    \begin{subfigure}{0.345\linewidth}
        \includegraphics[width=\linewidth]{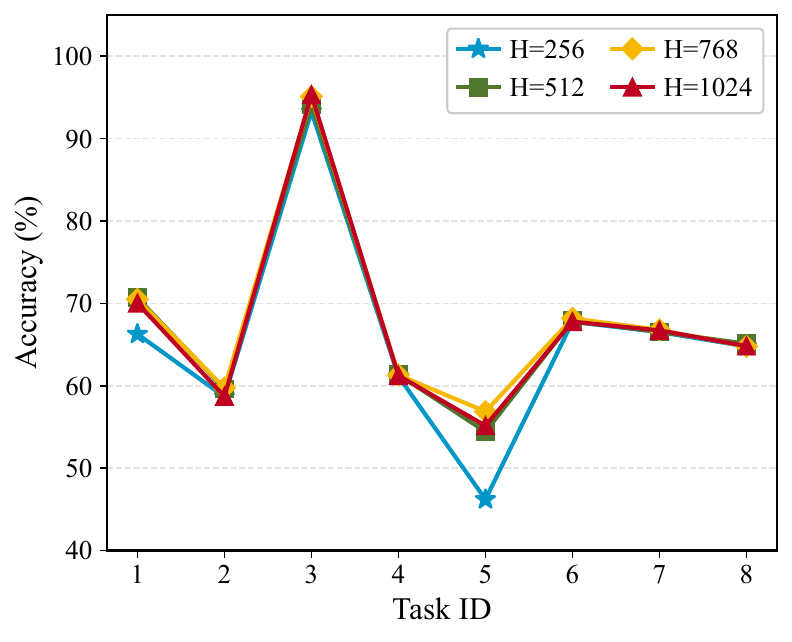}
    \end{subfigure}
    \caption{    
Routing ablation and generator hidden-dimension sensitivity on the CoIN benchmark. Left: comparison among oracle routing, learned prototype routing, and a no-routing variant that always uses the last-task generator. Right: per-task final accuracy under different generator hidden dimensions $H$.
    }
    \label{fig:rout_h}
\end{figure}

\begin{figure*}[t]
	\centering
	       \begin{minipage}{\textwidth}
        \centering
        \includegraphics[width=1\textwidth]{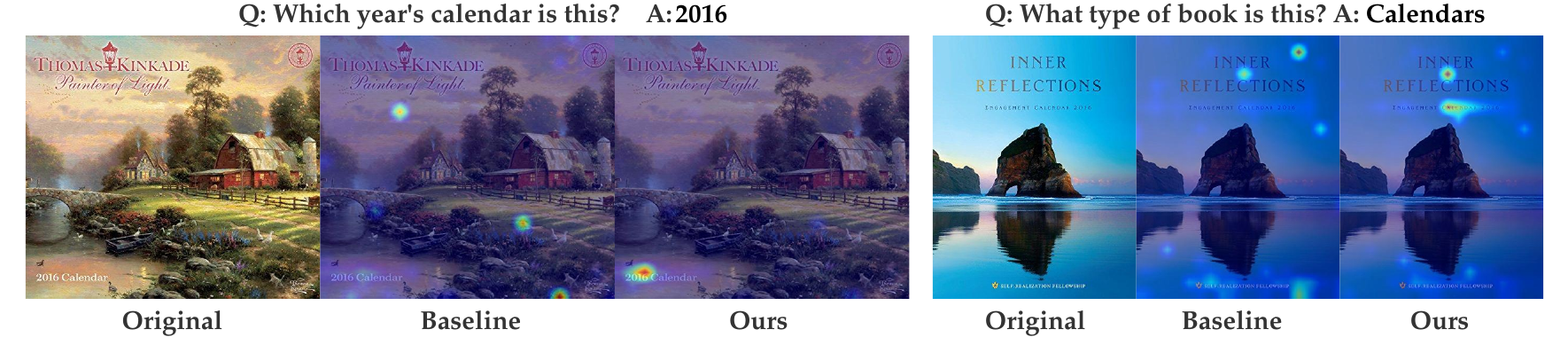} \\
        \label{fig:input}
    \end{minipage}
\caption{
Prompt-to-image attention visualizations on OCR-VQA examples.
For each example, \textbf{Column 1} shows the original image, \textbf{Column 2} shows the baseline attention map, and \textbf{Column 3} shows the attention map produced by \mame.
By generating prompts conditioned on both textual instructions and visual features, \name attends more selectively to query-relevant textual evidence, such as the printed year in the left example and the calendar-related text in the right example.
}
	\label{fig:attention-vis}
\end{figure*}

\noindent{\bf Routing Behavior and Generator Hidden Dimension.} Figure~\ref{fig:rout_h} presents two additional analyses. 
The left panel shows that our learned prototype router achieves an average accuracy of 67.48, closely approaching the oracle upper bound (68.17) and vastly outperforming the no-routing baseline (57.65). This advantage is most prominent on ScienceQA and ImageNet, where the router effectively prevents the severe task mismatches seen in the baseline. To further illustrate this, we provide case studies in Appendix~\ref{app:case-study}, showing that the advantage of \name is especially clear on samples requiring fine-grained instance-level adaptation. The right panel evaluates the generator's hidden dimension ($H$). While $H=256$ underperforms (especially on VizWiz), increasing $H$ from 512 to 768 or 1024 yields only marginal, plateauing improvements. Overall, the prototype router proves highly effective, and a moderate capacity suffices; thus, we adopt $H=512$ as our default setting.

\noindent \textbf{Visualization.}
Figure~\ref{fig:attention-vis} visualizes prompt-to-image attention maps on OCR-VQA examples, where \textit{Baseline} denotes the static task-level prompt variant. By generating prompts conditioned on both textual instructions and visual features, \name attends more selectively to query-relevant textual regions. For the year-recognition example, \name focuses on the lower-left text containing \textit{``2016 Calendar''}; for the book-type example, it attends to \textit{``Engagement Calendar 2016''}, which supports the answer \textit{``Calendars''}. Compared with the more diffuse baseline attention, these results suggest that cross-modal prompt generation helps \name capture visual evidence required by the current instruction. Additional qualitative case studies are provided in  Appendix~\ref{app:case-study}.

\section{Conclusion}
\label{sec:conclusion}
In this paper, we study multimodal continual instruction tuning with a particular focus on how models adapt to diverse samples encountered over a task stream. We find that continual adaptation should operate not only across tasks, but also across distinct samples within the same task, since different query-image pairs may require substantially different grounding, reasoning, or text-understanding behaviors. To address this challenge, we propose \mame, a framework that replaces static task-level adaptation with instance-specific prompt generation conditioned on the current multimodal input. Extensive experiments on the CoIN and UCIT benchmarks demonstrate that \name achieves strong performance across diverse multimodal tasks and maintains competitive continual learning behavior.

\noindent\textbf{Limitations.}
While \name is developed and validated in the prompt-tuning setting considered in this paper, we have not yet extended the same instance-specific adaptation idea to other lightweight adaptation paradigms such as LoRA. Exploring how instance-specific generation can be integrated with LoRA-style updates is an interesting direction for future work.

\section*{Acknowledgments}

This work was supported in part by the Basic Research Program of Jiangsu (BK20251251), NSFC (62506160),  JSTJ-2025-147, Fundamental and Interdisciplinary Disciplines Breakthrough Plan of the Ministry of Education of China (No. JYB2025XDXM118), the 111 Center (No. B26023), and the Collaborative Innovation Center of Novel Software Technology and Industrialization. The authors declare no competing interests.

{
    \small
    \bibliographystyle{ieeenat_fullname}
    \bibliography{main}
}

\clearpage
\appendix
\section*{\centering Appendix}
\renewcommand{\thesection}{\Alph{section}}
\setcounter{section}{0}

In this appendix, we provide additional details and supplementary analyses for \mame, including evaluation metrics, null-space projection analysis, additional benchmark results, sensitivity analysis, efficiency comparison, routing diagnostics, qualitative case studies, descriptions of compared methods, detailed result matrices, and broader-impact discussion.

\noindent\textbf{Section~\ref{app:evaluation-metrics}} provides the definitions of the continual evaluation metrics used in our experiments, including backward transfer and mean accuracy.

\noindent\textbf{Section~\ref{app:nullspace-analysis}} analyzes why null-space gradient projection can reduce interference with previous tasks.

\noindent\textbf{Section~\ref{app:continuous-performance}} reports stage-wise continual performance on CoIN and UCIT, analyzing forgetting behavior throughout sequential training.

\noindent\textbf{Section~\ref{app:additional_coin_baselines}} reports additional CoIN benchmark comparisons with recent MCIT baselines.

\noindent\textbf{Section~\ref{app:prompt-length}} analyzes the sensitivity of different methods to the number of prompts or LoRA experts.

\noindent\textbf{Section~\ref{app:efficiency}} analyzes the training efficiency, storage overhead, and inference cost of \name compared with representative baselines.

\noindent\textbf{Section~\ref{app:routing}} presents routing diagnostics, including the row-normalized routing confusion matrix.

\noindent\textbf{Section~\ref{app:case-study}} provides qualitative case studies on GQA, VQAv2, and OCR-VQA to illustrate instance-specific prompt behavior.

\noindent\textbf{Section~\ref{app:comparing-method}} summarizes the compared baseline methods.

\noindent\textbf{Section~\ref{app:pseudocode}} summarizes the training and inference procedures of \mame.

\noindent\textbf{Section~\ref{app:full-results}} reports the detailed continual-learning result matrices on CoIN and UCIT.

\noindent\textbf{Section~\ref{app:broader-impacts}} discusses the broader impacts.

\section{Details of Evaluation Metrics}
\label{app:evaluation-metrics}

We provide the detailed definitions of the continual evaluation metrics used in our experiments. Let $\mathcal{A}_{s,t}$ denote the performance on task $s$ after training up to task $t$, where $T$ is the total number of tasks.

\noindent \textbf{Backward Transfer (BWT).}
BWT measures the performance drop on previously learned tasks and quantifies catastrophic forgetting at incremental stage $t$, defined as $B_t=\frac{1}{t-1}\sum_{s=1}^{t-1}\left(\mathcal{A}_{s,s}-\mathcal{A}_{s,t}\right)$ for $t=2,\dots,T$. We adopt a forgetting-oriented sign convention for backward transfer, where lower values indicate less forgetting and negative values indicate positive backward transfer. A lower BWT indicates better resistance to catastrophic forgetting.

\noindent \textbf{Mean Accuracy (MA).}
MA evaluates the average performance over all learned tasks at incremental stage $t$, defined as $M_t=\frac{1}{t}\sum_{s=1}^{t}\mathcal{A}_{s,t}$ for $t=1,\dots,T$. A higher MA indicates better continual learning ability throughout the task sequence.

For stage-wise continual evaluation, we report both metrics across incremental stages and further average them over $t=2,\dots,T$.

\section{Analysis of Null-Space Gradient Projection}
\label{app:nullspace-analysis}

We provide a first-order analysis to illustrate why the null-space gradient projection can reduce interference on previously learned tasks. We follow the notation in Section~\ref{subsec:nullspace} and make the layer index explicit: $\mathbf{M}^{(t)}_\ell$, $V_{\ell,\perp}$, and $\boldsymbol{\Pi}^{(t)}_\ell$ denote the second-moment matrix, complementary basis, and projection matrix for projector layer $\ell$, respectively. The same argument applies independently to each linear layer of the shared projector $\pi(\cdot)$.

Let $\tilde{\mathbf{v}}^{(\ell)} \in \mathbb{R}^{d_\ell}$ denote the input feature to layer $\ell$. After observing tasks up to $\mathcal{D}_t$, the empirical second-moment matrix maintained for this layer is
\begin{equation}
    \mathbf{M}_{\ell}^{(t)}
    =
    \hat{\mathbb{E}}_{\leq t}
    \left[
    \tilde{\mathbf{v}}^{(\ell)}
    \tilde{\mathbf{v}}^{(\ell)\top}
    \right],
\end{equation}
where $\hat{\mathbb{E}}_{\leq t}$ denotes the empirical average over features collected from tasks up to $t$. Its singular value decomposition is
\begin{equation}
    \mathbf{M}_{\ell}^{(t)}
    =
    U_{\ell}\Sigma_{\ell}V_{\ell}^{\top},
    \quad
    \Sigma_{\ell}
    =
    \mathrm{diag}(\sigma_{\ell,1},\dots,\sigma_{\ell,d_\ell}),
\end{equation}
with $\sigma_{\ell,1}\geq\cdots\geq\sigma_{\ell,d_\ell}\geq0$. Following the energy criterion in Section~\ref{subsec:nullspace}, we split the right singular vectors into the retained principal subspace and its complement:
\begin{equation}
    V_{\ell}
    =
    [V_{\ell,\parallel}, V_{\ell,\perp}],
\end{equation}
where $V_{\ell,\parallel}$ contains the top $r_\ell$ directions and $V_{\ell,\perp}$ contains the remaining directions. Let $r_{\ell,\epsilon}$ denote the smallest rank satisfying the energy-retention criterion. In practice, we set $r_\ell=\min(r_{\ell,\epsilon},d_\ell-1)$ to preserve at least one update direction. With $\epsilon=0.99$ and $d_\ell\ge100$ in our setting, this cap does not affect the energy-retention criterion. Therefore, we have
\begin{equation}
    \frac{\sum_{k=1}^{r_\ell}\sigma_{\ell,k}}
    {\sum_{k=1}^{d_\ell}\sigma_{\ell,k}}
    \geq \epsilon,
\end{equation}
which implies
\begin{equation}
    \sum_{k=r_\ell+1}^{d_\ell}\sigma_{\ell,k}
    \leq
    (1-\epsilon)
    \sum_{k=1}^{d_\ell}\sigma_{\ell,k}.
\end{equation}
Therefore, the empirical energy of previous-task features in the complementary subspace is bounded by
\begin{equation}
    \hat{\mathbb{E}}_{\leq t}
    \left[
    \|V_{\ell,\perp}^{\top}\tilde{\mathbf{v}}^{(\ell)}\|_2^2
    \right]
    =
    \mathrm{tr}
    \left(
    V_{\ell,\perp}^{\top}
    \mathbf{M}_{\ell}^{(t)}
    V_{\ell,\perp}
    \right)
    =
    \sum_{k=r_\ell+1}^{d_\ell}\sigma_{\ell,k}
    \leq
    (1-\epsilon)\mathrm{tr}(\mathbf{M}_{\ell}^{(t)}).
\end{equation}

In practice, we set $\epsilon=0.99$, which retains most of the empirical feature energy from previous tasks. As a result, previous-task features are expected to have only small components in the complementary subspace. Intuitively, for an earlier-task feature $\tilde{\mathbf{v}}_{\mathrm{old}}^{(\ell)}$, this means
\begin{equation}
    V_{\ell,\perp}^{\top}
    \tilde{\mathbf{v}}_{\mathrm{old}}^{(\ell)}
    \approx \mathbf{0}.
\end{equation}

Now consider the weight matrix $W_\ell$ of layer $\ell$. During training on the next task, the gradient is projected as
\begin{equation}
    \nabla W_\ell'
    =
    \nabla W_\ell
    \boldsymbol{\Pi}_{\ell}^{(t)}
    =
    \nabla W_\ell
    V_{\ell,\perp}V_{\ell,\perp}^{\top}.
\end{equation}
For an earlier-task feature $\tilde{\mathbf{v}}_{\mathrm{old}}^{(\ell)}$, the first-order change in the layer output induced by this projected update is
\begin{equation}
    \Delta \mathbf{o}_{\mathrm{old}}^{(\ell)}
    =
    -\eta
    \nabla W_\ell
    V_{\ell,\perp}V_{\ell,\perp}^{\top}
    \tilde{\mathbf{v}}_{\mathrm{old}}^{(\ell)},
\end{equation}
where $\eta$ is the learning rate. Since the projected update acts only through the complementary component of the old feature, we can rewrite it as
\begin{equation}
    \Delta \mathbf{o}_{\mathrm{old}}^{(\ell)}
    =
    -\eta
    \nabla W_\ell
    V_{\ell,\perp}
    \left(
    V_{\ell,\perp}^{\top}
    \tilde{\mathbf{v}}_{\mathrm{old}}^{(\ell)}
    \right)
    \approx \mathbf{0}.
\end{equation}
Equivalently, its magnitude can be bounded as
\begin{equation}
    \|\Delta \mathbf{o}_{\mathrm{old}}^{(\ell)}\|_2
    \leq
    \eta
    \|\nabla W_\ell\|_2
    \left\|
    V_{\ell,\perp}^{\top}
    \tilde{\mathbf{v}}_{\mathrm{old}}^{(\ell)}
    \right\|_2.
\end{equation}
Thus, when earlier-task features are well captured by the retained principal subspace, their components in $V_{\ell,\perp}$ are small, and the projected update has a limited first-order effect on their layer outputs.

In the ideal case where $\tilde{\mathbf{v}}_{\mathrm{old}}^{(\ell)} \in \mathrm{span}(V_{\ell,\parallel})$, we have
\begin{equation}
    V_{\ell,\perp}^{\top}
    \tilde{\mathbf{v}}_{\mathrm{old}}^{(\ell)}
    =
    \mathbf{0},
\end{equation}
and therefore the projected update does not change the layer output for that feature to first order. Overall, this analysis shows why null-space projection helps reduce first-order interference on earlier-task representations while preserving update directions for new-task adaptation.

\subsection{Stability--Plasticity Analysis of Null-Space Projection}
\label{app:nsp-analysis}

Null-space gradient projection (NSP) is designed to preserve historical
multimodal representations by restricting updates along important feature
directions. However, such constraints may potentially reduce the model's
plasticity when learning new tasks. To analyze the trade-off between stability
and plasticity, we compare \name with a variant without NSP in terms of
immediate learning performance and final performance after completing the whole
task sequence. Specifically, the \emph{Immediate Avg. Accuracy} denotes the
average accuracy measured immediately after learning each task, which reflects
the model's ability to acquire new knowledge. The \emph{Final Avg. Accuracy}
denotes the average accuracy measured after completing the entire task
sequence, which reflects the retention of previously learned knowledge.

\begin{table}[t]
    \centering
    \setlength{\tabcolsep}{8pt}
    \renewcommand{\arraystretch}{1.15}
    \begin{tabular}{lcc}
    \toprule[1.2pt]
    Method & Immediate Avg. Accuracy & Final Avg. Accuracy\\
    \midrule
    w/o Null-Space Projection & 69.01 & 64.86\\
    \name & 68.05 & 67.48\\
    \bottomrule[1.2pt]
    \end{tabular}
    \caption{
    Stability--plasticity analysis of null-space projection on the CoIN
    benchmark. Immediate accuracy is measured after learning each task, while
    final accuracy is measured after completing the entire task sequence.
    }
    \label{tab:nsp-analysis}
\end{table}

As shown in Table~\ref{tab:nsp-analysis}, removing NSP improves the immediate
average accuracy from 68.05\% to 69.01\%, indicating that NSP introduces only a
minor restriction on learning new tasks. However, without NSP, the final
average accuracy decreases from 67.48\% to 64.86\%, demonstrating increased
interference from subsequent task updates. These results show that NSP achieves
a favorable stability--plasticity trade-off by preserving sufficient adaptation
capacity for new tasks while improving knowledge retention throughout
continual learning.

\section{Continual Learning Performance Across Tasks}
\label{app:continuous-performance}

We report stage-wise continual metrics on CoIN and UCIT to further analyze forgetting behavior during sequential training. Table~\ref{tab:continuous-metrics} shows the results on CoIN. Compared with previous methods, \name achieves the lowest average backward-transfer value and the highest average mean accuracy among the compared methods, with $B=\textbf{1.50}$ and $M=\textbf{68.92}$. Relative to the strongest baseline ModalPrompt, this corresponds to 2.22 points lower backward transfer and 12.82 points higher mean accuracy. These results indicate that \name improves forgetting resistance while maintaining strong overall continual performance throughout the task stream.

Table~\ref{tab:ucit-continuous-metrics} further reports the stage-wise continual metrics on UCIT. \name again achieves the lowest average backward transfer, with $B=\textbf{0.04}$, slightly outperforming ModalPrompt (0.06) and showing larger improvements over the remaining baselines. Its $B_t$ values remain close to zero across incremental stages and even become negative at early stages, indicating limited forgetting and slight positive backward transfer. Meanwhile, \name obtains an average mean accuracy of \underline{\textbf{76.31}}, which is comparable to the best-performing baseline SEFE (\textbf{76.35}) and higher than the other compared methods. These results further show that \name preserves previously acquired knowledge effectively while retaining competitive adaptation to newly introduced tasks.
\begin{table*}[ht]
    \centering
    \setlength{\tabcolsep}{5pt}
    \renewcommand\arraystretch{1.3}
    \resizebox{\linewidth}{!}{
    \begin{tabular}{l |cc cc cc cc cc cc cc |cc}
    \toprule[1.3pt]
        \multirow{2}{*}{Method} &\multicolumn{2}{c}{TextVQA} &\multicolumn{2}{c}{ImageNet}& \multicolumn{2}{c}{GQA} &\multicolumn{2}{c}{VizWiz}& \multicolumn{2}{c}{Grounding}&\multicolumn{2}{c}{VQAv2} & \multicolumn{2}{c}{OCR-VQA} & \multicolumn{2}{c}{\textit{Average}}\\
        \cline{2-17}
        &$B_2 \downarrow$ & $M_2 \uparrow$ & $B_3 \downarrow$ & $M_3 \uparrow$ & $B_4 \downarrow$ & $M_4 \uparrow$ & $B_5 \downarrow$ & $M_5 \uparrow$ & $B_6 \downarrow$ & $M_6 \uparrow$ & $B_7 \downarrow$ & $M_7 \uparrow$ & $B_8 \downarrow$ & $M_8 \uparrow$ &  $B \downarrow$ &  $M \uparrow$\\
        \midrule
         Finetune & 44.30&44.14&65.53&32.52&64.42&22.75&51.98&25.55&67.08&5.71&43.89&23.65&35.61&29.06&53.26&26.20\\
         CODA-Prompt& 11.54&57.88&27.70&34.05&14.38&43.44&12.78&42.76&12.72&39.28&14.05&39.42&7.27&45.46&14.34&43.18\\
         MoELoRA &41.31&43.13&52.47&34.08&32.76&41.71&33.81&37.71&41.41&25.59&30.80&34.34&26.12&37.48&36.95&36.29\\
        ModalPrompt &6.55&64.50&4.40&56.34&3.16&57.63&4.51&54.15&3.98&50.96&2.02&54.07&1.41&55.06&3.72&56.10\\
        ProgLoRA &17.21&59.92&6.78&73.60&17.60&60.71&12.27&61.35&18.31&50.60&9.56&57.66&7.53&59.09&12.75&60.42\\
         \rowcolor{gray!10}{\name (Ours)}& \textbf{3.37}&\textbf{65.11}&\textbf{2.48}&\textbf{74.37}&\textbf{1.32}&\textbf{71.51}&\textbf{1.07}&\textbf{68.07}&\textbf{0.83}&\textbf{68.08}&\textbf{0.78}&\textbf{67.80}&\textbf{0.66}&\textbf{67.48}&\textbf{1.50}&\textbf{68.92}\\
         \bottomrule[1.3pt]
    \end{tabular}}
    \vspace{-2mm}
    \caption{Continual performance metrics at each incremental stage on the CoIN benchmark. $B_t$ and $M_t$ stand for \textit{Backward Transfer} and \textit{Mean Accuracy} at incremental stage $t$.}
    \label{tab:continuous-metrics}
\end{table*}

\begin{table*}[ht]
    \centering
    \setlength{\tabcolsep}{10pt}
    \renewcommand\arraystretch{1.3}
    \resizebox{\linewidth}{!}{
    \begin{tabular}{l |cc cc cc cc cc |cc}
    \toprule[1.3pt]
        \multirow{2}{*}{Method} & \multicolumn{2}{c}{ArxivQA} & \multicolumn{2}{c}{VizWiz} & \multicolumn{2}{c}{IconQA} & \multicolumn{2}{c}{CLEVR} & \multicolumn{2}{c}{Flickr30k} & \multicolumn{2}{c}{\textit{Average}} \\
        \cline{2-13}
        & $B_2 \downarrow$ & $M_2 \uparrow$ & $B_3 \downarrow$ & $M_3 \uparrow$ & $B_4 \downarrow$ & $M_4 \uparrow$ & $B_5 \downarrow$ & $M_5 \uparrow$ & $B_6 \downarrow$ & $M_6 \uparrow$ & $B \downarrow$ & $M \uparrow$ \\
        \midrule
        LoRA-FT     & 1.37 & 90.83 & 8.81 & 75.77 & 13.55 & 71.04 & 15.28 & 68.40 & 18.78 & 61.24 & 11.56 & 73.46 \\
        O-LoRA      & 1.33 & \textbf{92.40} & 6.47 & \textbf{78.26} & 11.18 & 72.70 & 13.20 & 69.02 & 13.99 & 64.35 & 9.23 & 75.35 \\
        MoELoRA     & 1.16 & 91.44 & 5.85 & 77.52 & 11.78 & 65.66 & 10.31 & 65.68 & 14.63 & 58.98 & 8.74 & 71.85 \\
        ModalPrompt & 0.07 & 69.20 & 0.00 & 62.20 & 0.04 & 56.45 & 0.04 & 54.57 & \textbf{0.15} & 52.53 & 0.06 & 58.99 \\
        CL-MoE      & 3.90 & 89.84 & 8.40 & 75.65 & 13.04 & 64.74 & 11.39 & 65.15 & 15.56 & 58.49 & 10.46 & 70.77 \\
        HiDe        & 2.04 & 90.55 & 1.53 & 78.18 & 4.80 & 74.08 & 6.44 & 67.94 & 9.20 & 62.29 & 4.80 & 74.61 \\
        SEFE        & 0.57 & 90.73 & 4.35 & 78.19 & 8.51 & \textbf{74.17} & 9.30 & 72.15 & 11.33 & 66.54 & 6.81 & \textbf{76.35} \\
        \rowcolor{gray!10}{\name (Ours)} & \textbf{-0.14} & 88.58 & \textbf{-0.02} & 77.59 & \textbf{0.02} & 73.82 & \textbf{0.17} & \textbf{72.16} & {0.16} & \textbf{69.41} & \textbf{0.04} & {76.31} \\
        \bottomrule[1.3pt]
    \end{tabular}}
    \caption{Continual performance metrics at each incremental stage on the UCIT benchmark. $B_t$ and $M_t$ denote the backward transfer and mean accuracy at incremental stage $t$.}
    \label{tab:ucit-continuous-metrics}
\end{table*}

\section{Additional CoIN Baseline Comparison}
\label{app:additional_coin_baselines}

To provide a more comprehensive comparison on the CoIN benchmark, we include
additional results of recent MCIT baselines under the CoIN setting. All task-wise
scores correspond to the final accuracy after completing the entire task
sequence. As shown in Table~\ref{tab:additional-coin-baselines}, \name{} achieves
the best average performance among the compared methods.

\begin{table*}[t]
    \centering
    \small
    \setlength{\tabcolsep}{4pt}
    \renewcommand{\arraystretch}{1.05}
    \resizebox{\linewidth}{!}{%
    \begin{NiceTabular}{@{}lccccccccc@{}}
        \toprule
        Method
        & ScienceQA
        & TextVQA
        & ImageNet
        & GQA
        & VizWiz
        & Grounding
        & VQAv2
        & OCR-VQA
        & Average
        \\
        \midrule

        ProgLoRA~\cite{yu2025progressive}
        & 74.84
        & 51.83
        & 83.90
        & 49.93
        & 53.87
        & 31.19
        & 62.71
        & 64.44
        & 59.09
        \\

        SEFE~\cite{chen2025sefe}
        & 75.35
        & 58.66
        & 83.10
        & 54.25
        & 48.85
        & 16.75
        & 65.35
        & 66.25
        & 58.57
        \\

        HiDe~\cite{guo2025hide}
        & 73.20
        & 56.92
        & 69.28
        & 61.33
        & 50.76
        & 59.18
        & 67.12
        & 64.76
        & 62.82
        \\

        O-LoRA~\cite{wang2023orthogonal}
        & 86.63
        & 48.78
        & 12.38
        & 61.67
        & 46.68
        & 66.33
        & 67.05
        & 55.58
        & 56.65
        \\

        CL-MoE~\cite{huai2025cl}
        & 81.37
        & 51.48
        & 22.95
        & 59.99
        & 39.01
        & 55.26
        & 67.60
        & 63.70
        & 55.17
        \\

        \rowcolor{gray!10}
        \name{} (Ours)
        & 70.67
        & 59.61
        & 94.16
        & 61.37
        & 54.43
        & 67.92
        & 66.53
        & 65.11
        & \textbf{67.48}
        \\

        \bottomrule
    \end{NiceTabular}}
    \caption{\small
    Additional comparison on the CoIN benchmark with LLaVA-v1.5-7B as the
    backbone. All task-wise scores correspond to the final accuracy after
    completing the entire task sequence. The best average accuracy is
    highlighted in bold.
    }
    \label{tab:additional-coin-baselines}
\end{table*}

\section{Prompt/LoRA Number Sensitivity}
\label{app:prompt-length}

\begin{figure}[t]
    \centering
    \begin{subfigure}{0.48\linewidth}
        \centering
        \includegraphics[width=\linewidth]{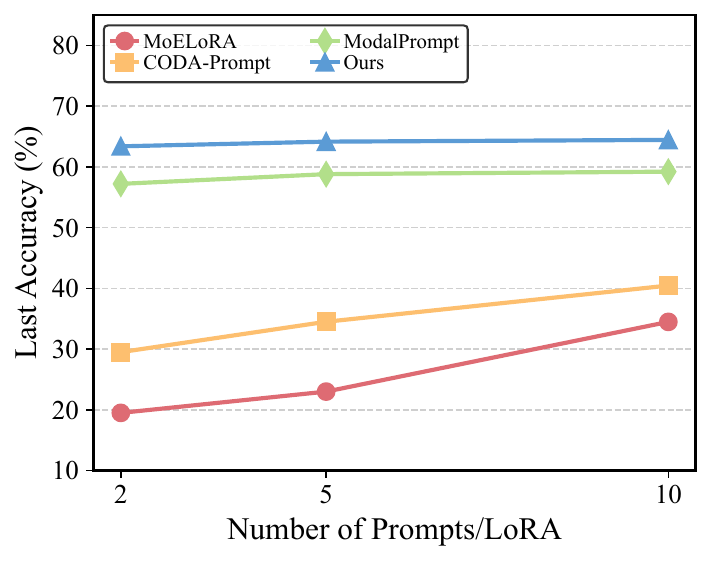}
    \end{subfigure}\hfill
    \begin{subfigure}{0.48\linewidth}
        \centering
        \includegraphics[width=\linewidth]{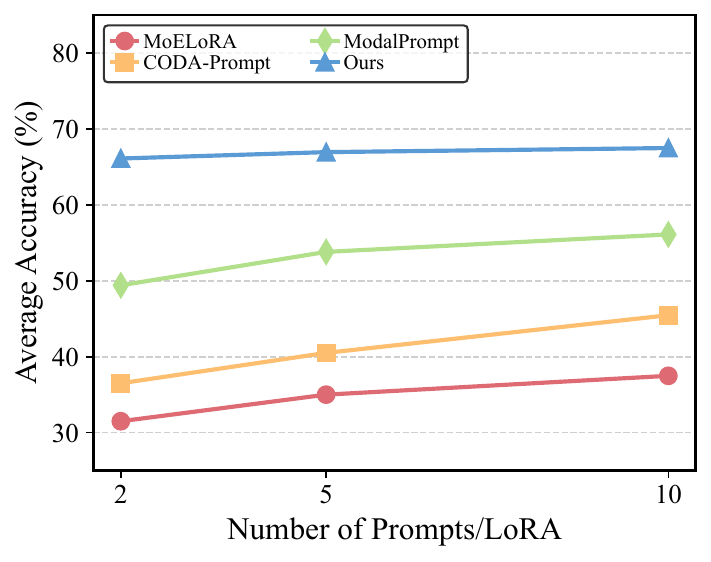}
    \end{subfigure}
    \caption{Impact of prompt and LoRA expert numbers on the CoIN benchmark. We vary the number of prompts for each task in prompt-based methods and the number of MoE LoRA experts in MoELoRA. Left: last-task accuracy. Right: final average accuracy.}
    \label{fig:prompt-length}
\end{figure}

We further analyze how the number of prompts or LoRA experts affects continual performance on the CoIN benchmark.
Figure~\ref{fig:prompt-length} illustrates the final average accuracy and the last-task accuracy when varying the
number of prompts for each task and the number of MoE LoRA experts. Specifically, the x-axis values $\{2,5,10\}$
denote the number of prompts used for each task in prompt-based methods and the number of LoRA experts in MoELoRA.
Generally, increasing the number of prompts or LoRA experts brings modest improvements, and the gains gradually
saturate once the number becomes moderately large. For baselines like MoELoRA, CODA-Prompt, and ModalPrompt,
increasing the corresponding number brings steady enhancements, whereas \name remains relatively stable across the
tested settings. Considering the trade-off between effectiveness and efficiency, we set the default number of prompts to $L_p=10$, which achieves strong performance within the evaluated prompt budget.

\section{Efficiency Comparison}
\label{app:efficiency}

We analyze the efficiency of \name from three aspects: training cost, storage
overhead, and inference efficiency.

\textbf{Training and storage overhead.}
Table~\ref{tab:efficiency-comparison} compares the training time and active
trainable parameters with representative prompt-based and LoRA-based continual
learning methods. Here, ``Trainable Param'' refers to the parameters updated
during a single task update, including the current task-specific generator and
the shared multimodal projector. \name updates 32M parameters per task and
achieves the lowest training time among all compared methods. For long-term
sequential learning, \name only stores one frozen task-specific generator for
each learned task, which occupies approximately 21.04 MB in our implementation.
Therefore, even after learning 50 tasks, the additional storage requirement is
about 1 GB, which is around one tenth of the full LLaVA backbone whose weights
alone require more than 10 GB. These results demonstrate that \name remains
lightweight for continual task expansion.

\textbf{Inference overhead.}
Since \name generates instance-specific prompts conditioned on each
image-instruction pair, we further evaluate the additional inference cost
introduced by dynamic prompt generation. The prompt generator is executed only
once before autoregressive decoding, after which \name follows the standard
autoregressive inference procedure with KV caching. Table~\ref{tab:inference-efficiency}
reports the prompt construction time, end-to-end latency, token throughput, and
peak memory usage.
To isolate the overhead introduced by instance-specific prompt generation, we include a static task-level prompt variant that uses the same backbone and training protocol but replaces input-conditioned prompt generation with learnable fixed task-level prompts. 
We also compare with representative MCIT baselines,
including SEFE and HiDe. Compared with static prompting, \name introduces only
1.78 ms additional prompt construction time and increases the end-to-end latency
by 2.50 ms per sample. The token throughput remains nearly unchanged,
decreasing from 30.55 to 30.21 tokens/s, while the peak memory increases by
only 0.30 GB. Compared with existing MCIT baselines, \name achieves a favorable
efficiency-performance trade-off by providing finer-grained multimodal
adaptation with limited additional computational and memory overhead.

\begin{table}[t]
    \centering
    \scriptsize

    \begin{minipage}[t]{0.38\linewidth}
    \centering
    \setlength{\tabcolsep}{5pt}
    \renewcommand\arraystretch{1.15}
    \begin{tabular}{lcc}
    \toprule[1.2pt]
        Method & Training (Hour) & Trainable Param\\
    \midrule
        MoELoRA & 7.27 & 360M \\
        CODA-Prompt & 3.43 & 165M \\
        HiDe & 3.93 & 247M \\
        ModalPrompt & 2.54 & 20M \\
        \name (Ours) & 1.59 & 32M \\
    \bottomrule[1.2pt]
    \end{tabular}
    \caption{Training efficiency comparison on CoIN.}
    \label{tab:efficiency-comparison}
    \end{minipage}
    \hfill
    \begin{minipage}[t]{0.59\linewidth}
    \centering
    \setlength{\tabcolsep}{3pt}
    \renewcommand\arraystretch{1.15}
    \begin{tabular}{lcccc}
    \toprule[1.2pt]
    Method & Prompt Construction & Latency & Throughput & Memory\\
    & (ms) & (ms/sample) & (Tokens/s) & (GB)\\
    \midrule
    Zero-shot
    & -- & 182.01 & 33.22 & 14.35\\
    SEFE
    & -- & 117.16 & 25.65 & 14.36\\
    HiDe
    & -- & 622.78 & 3.21 & 15.09\\
    Static Prompt
    & 0.40 & 195.49 & 30.55 & 14.94\\
    \name (Ours)
    & 2.18 & 197.99 & 30.21 & 15.24\\
    \bottomrule[1.2pt]
    \end{tabular}
    \caption{Inference overhead comparison on CoIN.}
    \label{tab:inference-efficiency}
    \end{minipage}
\end{table}

\section{Routing Diagnostics}
\label{app:routing}

\begin{figure}[t]
    \centering
    \includegraphics[width=0.92\linewidth]{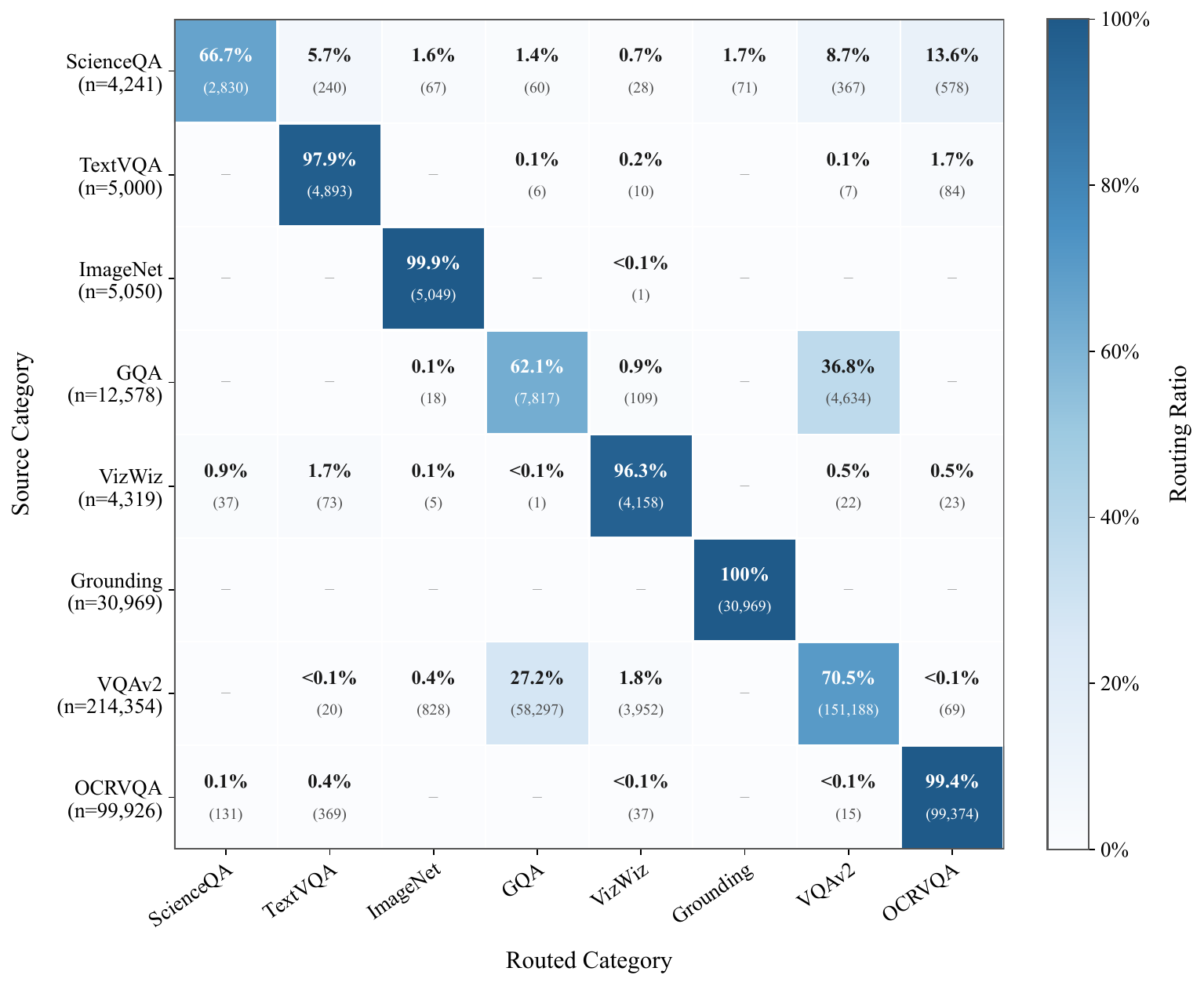}
\caption{Row-normalized routing confusion matrix on the final task. Each row corresponds to a ground-truth task and sums to 100\%. Diagonal entries indicate routing accuracy, while off-diagonal entries reveal the dominant misrouting patterns. The largest off-diagonal entries occur between GQA and VQAv2, and from ScienceQA to OCR-VQA and VQAv2.}
    \label{fig:routing-confusion}
\end{figure}

Figure~\ref{fig:routing-confusion} complements the routing ablation by showing where the learned router departs from oracle task selection. The dominant diagonal entries indicate that the router usually activates the correct task generator, which helps explain why learned routing remains close to the oracle-routing upper bound and far outperforms the no-routing variant in the main results. The largest off-diagonal errors reveal two main patterns. First, GQA and VQAv2 are strongly confused with each other, with GQA samples often routed to VQAv2 and VQAv2 samples often routed to GQA. Second, ScienceQA has a more dispersed routing pattern, with a noticeable fraction of samples routed to OCR-VQA and VQAv2. These errors do not necessarily lead to proportional drops in final task performance, since the misrouted generators often correspond to semantically related visual-question-answering or text-reading behaviors. This helps explain why the learned router remains close to oracle routing despite imperfect task-level routing accuracy. Because task sizes differ substantially, the row-normalized view is more informative than raw confusion counts for cross-task comparison.

\section{Case Studies}
\label{app:case-study}

To further examine the qualitative behavior of \name beyond aggregate accuracy, we present representative case studies
on GQA, VQAv2, and OCR-VQA. In all figures, \textit{Baseline} denotes the static task-level prompt variant, while
\textit{Drape} denotes our instance-specific prompt variant. Each example reports the prediction of the two variants
together with the ground-truth answer. The left example in each figure corresponds to a relatively simple case where both
methods succeed, while the middle and right examples illustrate more challenging cases in which the static prompt fails
but \name produces the correct answer. These examples provide qualitative evidence that a single shared task-level prompt
can be sufficient for standard inputs, but may become less reliable when the required adaptation depends on fine-grained
query--image interactions.

\begin{figure*}
  \centering
  \includegraphics[width=\textwidth]{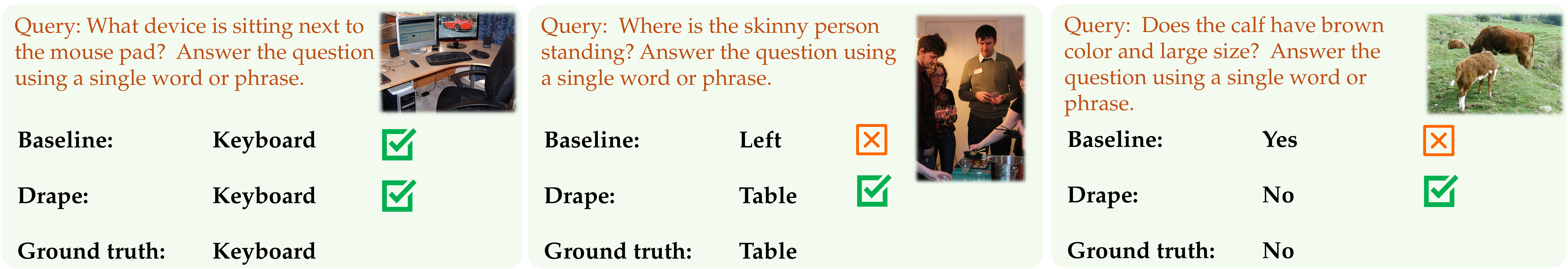}
\caption{Case studies on GQA. The left example is a relatively simple case where both variants are correct. The middle and right examples require finer spatial grounding and compositional attribute verification, respectively. In these more challenging cases, the static task-level prompt fails, whereas the instance-specific prompt generated by \name produces the correct answer.}
  \label{fig:case-gqa}
\end{figure*}

\noindent{\bf GQA.}
Figure~\ref{fig:case-gqa} shows three examples from GQA. In the left example, both variants correctly answer the query
\emph{``What device is sitting next to the mouse pad?''} with ``Keyboard,'' suggesting that both prompt strategies can
handle a direct object-grounding case. The middle example requires more precise spatial grounding: for the query
\emph{``Where is the skinny person standing?''}, the static prompt predicts the coarse spatial answer ``Left,'' whereas
\name correctly identifies the supporting object, ``Table.'' In the right example, the query
\emph{``Does the calf have brown color and large size?''} requires compositional attribute verification. The static
variant incorrectly predicts ``Yes,'' while \name correctly answers ``No.'' These cases suggest that instance-specific
prompt generation can better accommodate shifts from simple grounding to finer spatial reasoning and compositional
attribute judgment within the same task.

\begin{figure*}
\vspace{-4mm}
  \centering
  \includegraphics[width=\textwidth]{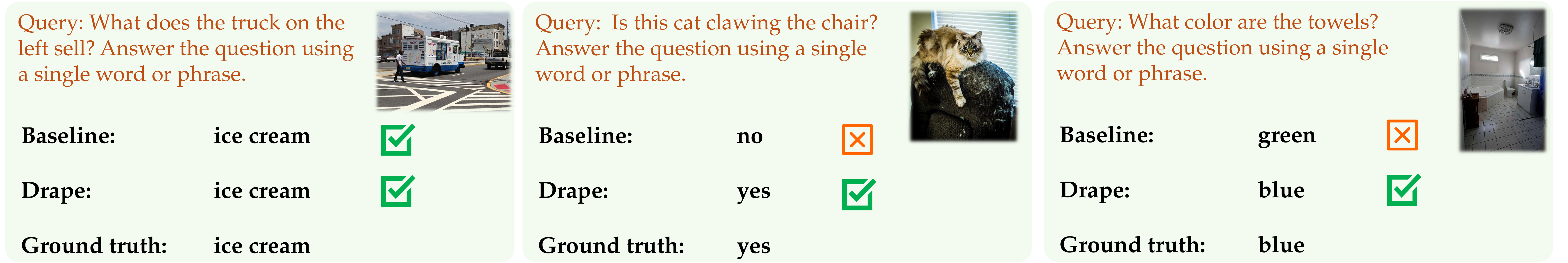}
\caption{Case studies on VQAv2. The left example is a relatively simple case where both variants succeed. The middle and right examples require fine-grained local action recognition and attribute recognition, respectively. In these cases, the static task-level prompt produces incorrect predictions, whereas \name remains aligned with the visual evidence relevant to the current query.}
  \label{fig:case-vqa}
\end{figure*}

\noindent{\bf VQAv2.}
Figure~\ref{fig:case-vqa} presents examples from VQAv2. In the left example, both variants correctly answer
\emph{``What does the truck on the left sell?''} with ``ice cream,'' indicating that a static task-level prompt can be
adequate for a straightforward open-domain VQA query. The middle example asks
\emph{``Is this cat clawing the chair?''}, which depends on fine-grained local action recognition. The static prompt
predicts ``no,'' whereas \name correctly predicts ``yes.'' In the right example, the query
\emph{``What color are the towels?''} requires local attribute recognition; the static prompt answers ``green,'' while
\name correctly answers ``blue.'' These examples indicate that the benefit of instance-specific prompting is not limited
to relational reasoning, but also extends to subtle local visual evidence in general VQA.

\begin{figure*}
  \centering
  \includegraphics[width=\textwidth]{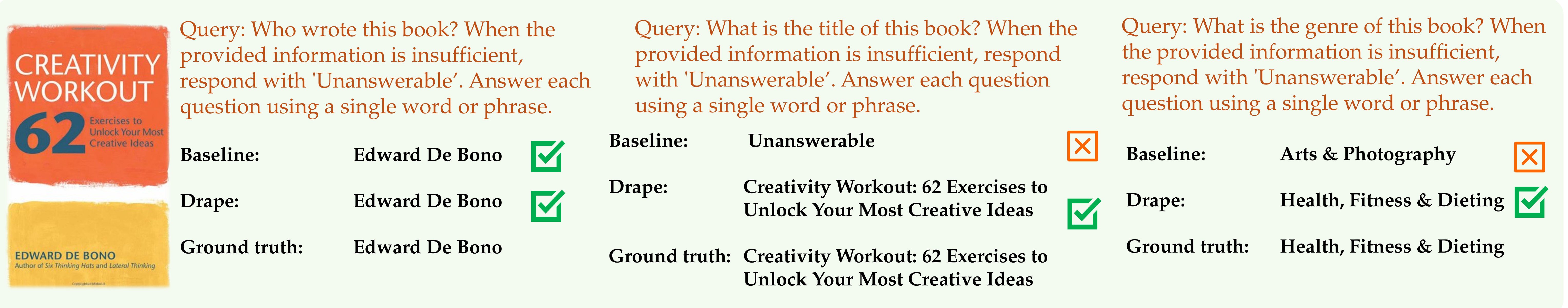}
\caption{OCR-VQA case study with a fixed image and different queries. Although the visual input is unchanged, the required behavior varies across queries, ranging from author recognition to title extraction and genre identification. The static task-level prompt handles the easier query but becomes less reliable on the more demanding ones, whereas \name adapts more effectively to the query-specific requirements.}
  \label{fig:case-ocr-image}
\end{figure*}

\noindent{\bf OCR-VQA: same image, different queries.}
Figure~\ref{fig:case-ocr-image} fixes the image and varies the query on the same book cover. This setting isolates the
effect of query-dependent adaptation, since the visual input remains unchanged while the required reasoning behavior
changes. In the left example, both variants correctly identify the author as ``Edward De Bono.'' In the middle example,
the query shifts to title extraction. The static prompt predicts ``Unanswerable,'' whereas \name recovers the full title
\emph{``Creativity Workout: 62 Exercises to Unlock Your Most Creative Ideas.''} In the right example, the query asks for
the book genre. The static prompt predicts the incorrect category \emph{``Arts \& Photography,''} while \name correctly
answers \emph{``Health, Fitness \& Dieting.''} These results suggest that even for a fixed image, different textual
queries may require different prompt behaviors, and that instance-specific prompt generation can better adapt to such
query-dependent semantic shifts.

\begin{figure}
  \centering
  \includegraphics[width=\textwidth]{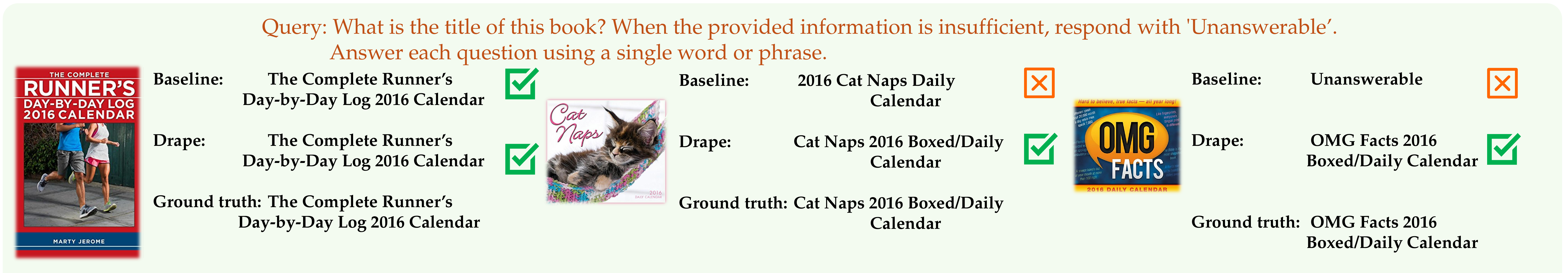}
\caption{OCR-VQA case study with a fixed query type and different images. Both variants correctly recover the title in a relatively simple example, while the more challenging covers require finer OCR-style extraction. In these harder cases, the static task-level prompt produces an incomplete answer or falls back to ``Unanswerable,'' whereas \name recovers the correct full title.}
  \label{fig:case-ocr-text}
\end{figure}

\noindent{\bf OCR-VQA: same query, different images.}
Figure~\ref{fig:case-ocr-text} provides the complementary analysis by fixing the query type and varying the image. All
examples ask for the book title. In the left example, both variants correctly recover the title from a relatively simple
cover. In the middle example, the static prompt predicts the incomplete title
\emph{``2016 Cat Naps Daily Calendar,''} while \name correctly recovers
\emph{``Cat Naps 2016 Boxed/Daily Calendar.''} In the right example, the static prompt predicts ``Unanswerable,''
whereas \name correctly identifies \emph{``OMG Facts 2016 Boxed/Daily Calendar.''} This comparison suggests that even
when the instruction form is fixed, the appropriate adaptation still depends on the specific visual layout and textual
content of each image.

\paragraph{Overall observation.}
Across GQA, VQAv2, and OCR-VQA, we observe a consistent qualitative pattern. Static task-level prompts are often
sufficient for relatively simple or canonical samples, but they can become brittle when the task requires fine-grained
grounding, local visual recognition, compositional attribute verification, or OCR-style text extraction. By generating
prompts conditioned on the current query--image pair, \name provides a more flexible adaptation mechanism that better
matches sample-specific multimodal demands. These case studies complement the quantitative results by illustrating how
instance-specific prompt generation can improve within-task adaptation in multimodal continual instruction tuning.

\section{Comparing Methods}
\label{app:comparing-method}

In this section, we provide brief descriptions of the methods compared in the paper.

\noindent \textbf{Finetune} sequentially fine-tunes the model on each incoming task without an explicit mechanism to
preserve previous knowledge.

\noindent \textbf{MoELoRA} maintains multiple task-related LoRA experts together with a gating function, so that only part of the additional
parameters are activated for each input. In this way, it aims to preserve previously acquired knowledge while adapting to
the current sample through expert selection.

\noindent \textbf{L2P} maintains a prompt pool in memory space and retrieves relevant prompts for each input, thereby balancing
task-invariant and task-specific knowledge through prompt selection.

\noindent \textbf{DualPrompt} introduces two types of prompts, namely general prompts and expert prompts, to model task-invariant and
task-specific knowledge, respectively. These prompts are attached to different transformer layers to facilitate continual
adaptation and knowledge retention.

\noindent \textbf{CODA-Prompt} learns a set of input-conditioned prompts for rehearsal-free continual learning and dynamically composes them during inference.

\noindent \textbf{ModalPrompt} extends prompt-based continual learning to the multimodal setting by introducing learnable prompts for vision-language adaptation while keeping most backbone parameters frozen.

\noindent \textbf{ProgLoRA} incrementally expands task-specific LoRA parameters across tasks and preserves previous knowledge by freezing historical modules. We follow the standard configuration in its released implementation and use the default
LoRA hyperparameters.

\noindent \textbf{LoRA-FT} performs sequential low-rank adaptation without an explicit forgetting-mitigation mechanism. 

\noindent \textbf{O-LoRA} reduces interference across tasks by imposing orthogonality constraints on LoRA updates. It aims
to preserve previously learned knowledge while maintaining adaptation ability on new tasks.

\noindent \textbf{CL-MoE} is an expert-based continual tuning method that improves adaptation through task-specialized
modules and expert routing. It dynamically selects or combines experts according to the current input.

\noindent \textbf{HiDe} is motivated by CKA similarity analysis, which suggests that top and lower transformer layers
exhibit different representation patterns in continual learning. It hierarchically decouples adaptation by using
task-specific LoRA expansion with dual-modality anchor matching in higher layers, while fusing LoRAs in lower layers to
preserve shared knowledge without router training.

\noindent \textbf{SEFE} addresses forgetting through answer-style diversification and RegLoRA. RegLoRA regularizes the
top-$M\%$ elements of LoRA update matrices to preserve critical historical knowledge during continual adaptation.

\begin{algorithm}[t]
\caption{Training of \name for Multimodal Continual Instruction Tuning}
\label{alg:drape-training}
\begin{algorithmic}[1]
\Require Task stream $\{\mathcal{D}_t\}_{t=1}^{T}$, model components $\phi,\psi,\theta,\pi,\{G_t\}_{t=1}^{T}$, CLIP encoders $\xi,\gamma$, energy threshold $\epsilon$
\Ensure Trained projector $\pi$, frozen generators $\{G_t\}_{t=1}^{T}$, task prototypes $\{\mathbf{c}_t\}_{t=1}^{T}$
\State Initialize cumulative feature statistics $\mathbf{M}_{\ell}^{(0)}=\mathbf{0}$ and counts $N_{\ell}^{(0)}=0$ for each projector layer $\ell$
\For{$t=1$ to $T$}
    \State Activate $G_t$ and freeze historical generators
    \For{each mini-batch $\mathcal{B}\subset\mathcal{D}_t$}
        \State Compute $\mathbf{w}_i=\pi(\phi(\mathbf{v}_i))$, $\mathbf{u}_i=\psi(\mathbf{q}_i)$, and $\mathbf{P}_i=G_t(\mathbf{w}_i,\mathbf{u}_i)$
        \State Accumulate projector-layer feature statistics and cache CLIP routing features $\tilde{\mathbf{e}}_i$
        \State Compute autoregressive loss $\mathcal{L}_t$ and backpropagate gradients
        \If{$t>1$}
            \State Project gradients of $\pi$: $\nabla W_\ell\leftarrow\nabla W_\ell\boldsymbol{\Pi}_{\ell}^{(t-1)}$
        \EndIf
        \State Update $G_t$ and $\pi$
    \EndFor
    \State Freeze $G_t$ and update cumulative statistics $\mathbf{M}_{\ell}^{(t)}$ and $N_{\ell}^{(t)}$
    \State Compute $\boldsymbol{\Pi}_{\ell}^{(t)}$ from $\mathbf{M}_{\ell}^{(t)}$ using SVD and threshold $\epsilon$
    \State Initialize $\mathbf{c}_t$ as the normalized mean of cached CLIP features
    \If{$t>1$}
        \State Refine $\mathbf{c}_t$ using Eq.~\eqref{eq:prototype_loss}
    \EndIf
    \State Register $\mathbf{c}_t$ and discard cached CLIP features
\EndFor
\end{algorithmic}
\end{algorithm}

\begin{algorithm}[t]
\caption{Inference of \name without Task Labels}
\label{alg:drape-inference}
\begin{algorithmic}[1]
\Require Test sample $(\mathbf{v},\mathbf{q})$, model components $\phi,\psi,\theta,\pi$, CLIP encoders $\xi,\gamma$, generators $\{G_t\}_{t=1}^{T}$, task prototypes $\{\mathbf{c}_t\}_{t=1}^{T}$
\Ensure Predicted answer $\hat{\mathbf{y}}$
\State Compute routing feature $\tilde{\mathbf{e}}=\mathrm{norm}([\xi(\mathbf{q});\gamma(\mathbf{v})])$
\State Select generator $t^*=\arg\max_{t\in\{1,\dots,T\}}\cos(\tilde{\mathbf{e}},\mathbf{c}_t)$
\State Compute visual features $\mathbf{w}=\pi(\phi(\mathbf{v}))$ and text embeddings $\mathbf{u}=\psi(\mathbf{q})$
\State Generate instance-specific prompt $\mathbf{P}=G_{t^*}(\mathbf{w},\mathbf{u})$
\State Generate prediction $\hat{\mathbf{y}}$ using $\theta$ conditioned on $[\mathbf{P};\mathbf{w};\mathbf{u}]$
\end{algorithmic}
\end{algorithm}

\section{Pseudocode of \textsc{Drape}}
\label{app:pseudocode}

We summarize the training and inference procedures of \name in Algorithms~\ref{alg:drape-training} and~\ref{alg:drape-inference}. During sequential training, \name learns a task-specific prompt generator while preserving previously learned generators. The shared multimodal projector is updated using layer-wise null-space gradient projection to reduce interference with earlier tasks. After each task, the cumulative feature statistics and projection matrices are updated, and a compact CLIP-space prototype is registered for task-free routing. During inference, the prototype router selects the most relevant frozen generator based on the current image-instruction pair, enabling instance-specific prompt generation without requiring task labels.

\section{Detailed Result Matrices}
  \label{app:full-results}
In Tables~\ref{tab:coin}–\ref{tab:ucit}, we report the final accuracy matrices of the compared methods on the CoIN and UCIT benchmarks. Results may vary across hardware and software environments.

\begin{table*}[!t]
\centering
\caption{Detailed continual instruction tuning results matrices for different methods on the CoIN benchmark.}
\label{tab:coin}

\begin{subtable}[t]{0.49\linewidth}
\centering
\resizebox{\linewidth}{!}{
\begin{tabular}{lcccccccc}
\toprule
\textbf{Finetune} & ScienceQA & TextVQA & ImageNet & GQA & VizWiz & Grounding & VQAv2 & OCR-VQA \\
\midrule
ScienceQA  & 82.45 \\
TextVQA    & 38.15 & 50.14 \\
ImageNet   & 0.96 & 0.58 & 96.03 \\
GQA        & 13.91 & 15.78 & 5.67 & 55.65 \\
VizWiz     & 8.46 & 25.17 & 4.60 & 38.12 & 51.42 \\
Grounding  & 0.00 & 0.00 & 0.00 & 0.27 & 0.00 & 34.00 \\
VQAv2      & 9.10 & 27.58 & 6.62 & 43.92 & 19.10 & 0.03 & 59.17 \\
OCR-VQA     & 26.00 & 25.38 & 28.51 & 33.07 & 26.52 & 0.10 & 40.00 & 52.92 \\
\bottomrule
\end{tabular}}
\end{subtable}\hfill
\begin{subtable}[t]{0.49\linewidth}
\centering
\resizebox{\linewidth}{!}{
\begin{tabular}{lcccccccc}
\toprule
\textbf{MoELoRA} & ScienceQA & TextVQA & ImageNet & GQA & VizWiz & Grounding & VQAv2 & OCR-VQA \\
\midrule
ScienceQA  & 75.78 \\
TextVQA    & 34.47 & 51.80 \\
ImageNet   & 22.61 & 0.04 & 79.60 \\
GQA        & 32.37 & 34.04 & 42.48 & 57.95 \\
VizWiz     & 45.32 & 38.13 & 2.63 & 43.80 & 58.70 \\
Grounding  & 58.76 & 9.08 & 5.64 & 31.87 & 11.45 & 36.77 \\
VQAv2      & 33.01 & 48.42 & 10.61 & 49.78 & 32.23 & 1.75 & 64.58 \\
OCR-VQA     & 47.34 & 32.91 & 38.73 & 37.15 & 42.48 & 0.97 & 42.77 & 57.50 \\
\bottomrule
\end{tabular}}
\end{subtable}

\begin{subtable}[t]{0.49\linewidth}
\centering
\resizebox{\linewidth}{!}{
\begin{tabular}{lcccccccc}
\toprule
\textbf{L2P} & ScienceQA & TextVQA & ImageNet & GQA & VizWiz & Grounding & VQAv2 & OCR-VQA \\
\midrule
ScienceQA  & 72.83 \\
TextVQA    & 68.07 & 57.16 \\
ImageNet   & 32.05 & 26.73 & 39.43 \\
GQA        & 47.53 & 46.02 & 18.03 & 60.47 \\
VizWiz     & 65.94 & 37.68 & 1.72 & 56.29 & 47.90 \\
Grounding  & 5.74 & 42.96 & 33.92 & 39.44 & 39.64 & 1.87 \\
VQAv2      & 32.57 & 48.65 & 7.41 & 47.32 & 34.52 & 8.17 & 59.40 \\
OCR-VQA     & 54.42 & 46.04 & 30.36 & 57.09 & 42.19 & 9.38 & 50.45 & 54.03 \\
\bottomrule
\end{tabular}}
\end{subtable}\hfill
\begin{subtable}[t]{0.49\linewidth}
\centering
\resizebox{\linewidth}{!}{
\begin{tabular}{lcccccccc}
\toprule
\textbf{Dualprompt} & ScienceQA & TextVQA & ImageNet & GQA & VizWiz & Grounding & VQAv2 & OCR-VQA \\
\midrule
ScienceQA  & 67.16 \\
TextVQA    & 52.20 & 53.12 \\
ImageNet   & 28.49 & 24.77 & 46.40 \\
GQA        & 49.70 & 47.94 & 12.06 & 55.10 \\
VizWiz     & 57.88 & 51.17 & 21.34 & 48.03 & 51.62 \\
Grounding  & 18.27 & 39.64 & 29.77 & 44.06 & 35.97 & 30.82 \\
VQAv2      & 36.77 & 49.85 & 22.48 & 27.96 & 41.08 & 13.51 & 61.27 \\
OCR-VQA     & 56.40 & 47.12 & 34.96 & 42.03 & 44.14 & 12.01 & 54.43 & 53.36 \\
\bottomrule
\end{tabular}}
\end{subtable}

\begin{subtable}[t]{0.49\linewidth}
\centering
\resizebox{\linewidth}{!}{
\begin{tabular}{lcccccccc}
\toprule
\textbf{CODA-Prompt} & ScienceQA & TextVQA & ImageNet & GQA & VizWiz & Grounding & VQAv2 & OCR-VQA \\
\midrule
ScienceQA  & 70.26 \\
TextVQA    & 58.72 & 57.05 \\
ImageNet   & 36.96 & 34.95 & 30.26 \\
GQA        & 50.78 & 53.52 & 10.12 & 59.35 \\
VizWiz     & 55.37 & 47.21 & 6.78 & 56.43 & 48.01 \\
Grounding  & 33.56 & 47.67 & 32.07 & 44.62 & 43.39 & 34.42 \\
VQAv2      & 48.34 & 49.54 & 20.72 & 47.72 & 35.73 & 13.03 & 60.87 \\
OCR-VQA     & 58.15 & 50.16 & 24.04 & 54.33 & 48.94 & 17.83 & 55.86 & 54.42 \\
\bottomrule
\end{tabular}}
\end{subtable}\hfill
\begin{subtable}[t]{0.49\linewidth}
\centering
\resizebox{\linewidth}{!}{
\begin{tabular}{lcccccccc}
\toprule
\textbf{ProgLoRA} & ScienceQA & TextVQA & ImageNet & GQA & VizWiz & Grounding & VQAv2 & OCR-VQA \\
\midrule
ScienceQA  & 76.27 \\
TextVQA    & 59.06 & 60.78 \\
ImageNet   & 70.92 & 52.57 & 97.32 \\
GQA        & 51.56 & 50.33 & 79.68 & 61.27 \\
VizWiz     & 65.62 & 50.79 & 81.17 & 48.99 & 60.16 \\
Grounding  & 40.08 & 47.53 & 77.75 & 50.61 & 48.30 & 39.35 \\
VQAv2      & 76.90 & 56.79 & 73.90 & 53.64 & 40.63 & 35.96 & 65.83 \\
OCR-VQA     & 74.84 & 51.83 & 83.90 & 49.93 & 53.87 & 31.19 & 62.71 & 64.44 \\
\bottomrule
\end{tabular}}
\end{subtable}

\begin{subtable}[t]{0.49\linewidth}
\centering
\resizebox{\linewidth}{!}{
\begin{tabular}{lcccccccc}
\toprule
\textbf{ModalPrompt} & ScienceQA & TextVQA & ImageNet & GQA & VizWiz & Grounding & VQAv2 & OCR-VQA \\
\midrule
ScienceQA  & 77.05 \\
TextVQA    & 70.50 & 58.50 \\
ImageNet   & 68.57 & 58.18 & 42.26 \\
GQA        & 68.82 & 56.08 & 43.43 & 62.17 \\
VizWiz     & 67.48 & 55.05 & 37.60 & 61.81 & 48.81 \\
Grounding  & 66.58 & 55.68 & 35.92 & 61.95 & 48.74 & 36.88 \\
VQAv2      & 68.12 & 56.43 & 40.22 & 60.92 & 51.19 & 36.63 & 64.99 \\
OCR-VQA     & 68.42 & 56.40 & 41.13 & 61.11 & 50.13 & 36.69 & 66.90 & 59.68 \\
\bottomrule
\end{tabular}}
\end{subtable}\hfill
\begin{subtable}[t]{0.49\linewidth}
\centering
\resizebox{\linewidth}{!}{
\begin{tabular}{lcccccccc}
\toprule
\textbf{\scshape{Drape}} & ScienceQA & TextVQA & ImageNet & GQA & VizWiz & Grounding & VQAv2 & OCR-VQA \\
\midrule
ScienceQA  & 73.66 \\
TextVQA    & 70.29 & 59.93 \\
ImageNet   & 69.21 & 59.43 & 94.46 \\
GQA        & 69.87 & 59.71 & 94.50 & 61.96 \\
VizWiz     & 69.79 & 59.54 & 94.53 & 61.88 & 54.60 \\
Grounding  & 70.01 & 59.51 & 94.46 & 61.93 & 54.55 & 68.03 \\
VQAv2      & 70.57 & 59.49 & 94.32 & 61.19 & 54.32 & 68.07 & 66.65 \\
OCR-VQA     & 70.67 & 59.61 & 94.16 & 61.37 & 54.43 & 67.92 & 66.53 & 65.11 \\
\bottomrule
\end{tabular}}
\end{subtable}

\end{table*}

\begin{table*}[!t]
\centering
\caption{Detailed continual instruction tuning results matrices for different methods on the UCIT benchmark.}
\label{tab:ucit}
\begin{subtable}[t]{0.49\linewidth}
\centering
\resizebox{\linewidth}{!}{
\begin{tabular}{lcccccc}
\toprule
\textbf{LoRA-FT} & ImgNet-R & ArxivQA & VizWiz & IconQA & CLEVR & Flickr30k \\
\midrule
ImgNet-R & 92.00 & & & & & \\
ArxivQA & 90.63 & 91.03 & & & & \\
VizWiz & 74.00 & 91.40 & 61.90 & & & \\
IconQA & 72.73 & 78.53 & 53.03 & 79.87 & & \\
CLEVR & 68.97 & 77.60 & 49.43 & 67.67 & 78.33 & \\
Flickr30k & 58.03 & 77.63 & 44.39 & 67.40 & 61.77 & 58.22 \\
\bottomrule
\end{tabular}}
\end{subtable}\hfill
\begin{subtable}[t]{0.49\linewidth}
\centering
\resizebox{\linewidth}{!}{
\begin{tabular}{lcccccc}
\toprule
\textbf{O-LoRA} & ImgNet-R & ArxivQA & VizWiz & IconQA & CLEVR & Flickr30k \\
\midrule
ImgNet-R & 91.40 & & & & & \\
ArxivQA & 90.07 & 94.73 & & & & \\
VizWiz & 81.50 & 91.70 & 61.59 & & & \\
IconQA & 82.40 & 77.73 & 54.06 & 76.63 & & \\
CLEVR & 81.07 & 77.87 & 50.62 & 62.00 & 73.53 & \\
Flickr30k & 77.50 & 78.07 & 44.50 & 63.13 & 64.73 & 58.16 \\
\bottomrule
\end{tabular}}
\end{subtable}
\begin{subtable}[t]{0.49\linewidth}
    \centering
    \resizebox{\linewidth}{!}{
    \begin{tabular}{lcccccc}
    \toprule
        \textbf{MoELoRA} & ImgNet-R & ArxivQA & VizWiz & IconQA & CLEVR & Flickr30k \\
    \midrule
       ImgNet-R & 91.23 &  &  &  &  &  \\
       ArxivQA & 90.07 & 92.80 &  &  &  &  \\
       VizWiz & 80.30 & 92.03 & 60.22 &  &  &  \\
       IconQA & 80.53 & 77.20 & 51.19 & 53.73 &  &  \\
       CLEVR & 79.37 & 77.40 & 48.68 & 51.30 & 71.67 & \\
       Flickr30k & 70.07 & 77.70 & 44.69 & 50.03 & 54.03 & 57.34  \\
    \bottomrule
    \end{tabular}}
\end{subtable}\hfill
\begin{subtable}[t]{0.49\linewidth}
\centering
\resizebox{\linewidth}{!}{
    \begin{tabular}{lcccccc}
    \toprule
       \textbf{ModalPrompt} & ImgNet-R & ArxivQA & VizWiz & IconQA & CLEVR & Flickr30k \\
    \midrule
       ImgNet-R & 51.10 &  &  &  &  &   \\
       ArxivQA & 51.03 & 87.37 &  &  &  &  \\
       VizWiz & 51.10 & 87.37 & 48.14 &  &  &  \\
       IconQA & 51.20 & 87.27 & 48.03 & 39.30 &  &  \\
       CLEVR & 51.17 & 87.23 & 47.96 & 39.40 & 47.07 & \\
       Flickr30k & 51.07 & 87.27 & 48.11 & 39.23 & 46.57 & 42.93  \\
    \bottomrule
    \end{tabular}}
\end{subtable}
\begin{subtable}[t]{0.49\linewidth}
    \centering
    \resizebox{\linewidth}{!}{
    \begin{tabular}{lcccccc}
    \toprule
       \textbf{CL-MoE} & ImgNet-R & ArxivQA & VizWiz & IconQA & CLEVR & Flickr30k \\
    \midrule
       ImgNet-R & 91.30 &  &  &  &  &   \\
       ArxivQA & 87.40 & 92.27 &  &  &  &  \\
       VizWiz & 76.23 & 90.53 & 60.18 &  &  &  \\
       IconQA & 77.33 & 76.20 & 51.09 & 54.33 &  &  \\
       CLEVR & 75.73 & 75.70 & 48.27 & 52.80 & 73.23 & \\
       Flickr30k & 66.33 & 77.00 & 44.78 & 51.87 & 53.53 & 57.42  \\
    \bottomrule
    \end{tabular}}
\end{subtable}\hfill
\begin{subtable}[t]{0.49\linewidth}
\centering
\resizebox{\linewidth}{!}{
    \begin{tabular}{lcccccc}
    \toprule
       \textbf{HiDe} & ImgNet-R & ArxivQA & VizWiz & IconQA & CLEVR & Flickr30k \\
    \midrule
       ImgNet-R & 90.87 &  &  &  &  &   \\
       ArxivQA & 88.83 & 92.27 &  &  &  &  \\
       VizWiz & 88.17 & 91.90 & 54.46 &  &  &  \\
       IconQA & 87.00 & 89.70 & 46.51 & 73.13 &  &  \\
       CLEVR & 86.23 & 90.07 & 44.28 & 64.37 & 54.77 & \\
       Flickr30k & 84.03 & 90.73 & 44.43 & 58.93 & 41.37 & 54.25  \\
    \bottomrule
    \end{tabular}}
\end{subtable}
\begin{subtable}[t]{0.49\linewidth}
    \centering
    \resizebox{\linewidth}{!}{
    \begin{tabular}{lcccccc}
    \toprule
       \textbf{SEFE} & ImgNet-R & ArxivQA & VizWiz & IconQA & CLEVR & Flickr30k \\
    \midrule
       ImgNet-R & 90.80 &  &  &  &  &   \\
       ArxivQA & 90.23 & 91.23 &  &  &  &  \\
       VizWiz & 83.60 & 89.73 & 61.24 &  &  &  \\
       IconQA & 84.17 & 77.20 & 56.36 & 78.93 &  &  \\
       CLEVR & 82.90 & 77.13 & 53.54 & 71.43 & 75.73 & \\
       Flickr30k & 80.83 & 78.00 & 47.01 & 69.63 & 65.83 & 57.92  \\
    \bottomrule
    \end{tabular}}
\end{subtable}\hfill
\begin{subtable}[t]{0.49\linewidth}
\centering
\resizebox{\linewidth}{!}{
    \begin{tabular}{lcccccc}
    \toprule
       \textbf{\scshape{Drape}} & ImgNet-R & ArxivQA & VizWiz & IconQA & CLEVR & Flickr30k \\
    \midrule
        ImgNet-R &85.13& &     &     &     &       \\
        ArxivQA &85.27&91.90& &     &     &       \\
        VizWiz &85.20&91.87&55.70& &     &       \\
        IconQA &85.27&91.63&55.77&62.63& &       \\
        CLEVR &85.23&91.57&55.62&62.27&66.10&   \\
        Flickr30k &85.07&91.60&55.68&62.53&65.77&55.82 \\
    \bottomrule
    \end{tabular}}
\end{subtable}
\end{table*}

\section{Broader Impacts}
\label{app:broader-impacts}

This work studies rehearsal-free and parameter-efficient multimodal continual instruction tuning, which may reduce the computational and storage costs of adapting MLLMs to evolving tasks while helping preserve previously learned capabilities. Such improvements can make continual updates more practical for research and resource-constrained applications. The broader risks are mainly those associated with MLLMs and continual adaptation in general: if newly introduced data contains biases, unsafe content, or systematic errors, the adapted model may inherit these issues, and more efficient adaptation could also facilitate unintended specialization. This work does not introduce new datasets, release high-risk models, or directly address deployment safety; practical use should therefore include careful data curation, safety evaluation, and post-update monitoring.
\clearpage
\section*{NeurIPS Paper Checklist}

\begin{enumerate}

\item {\bf Claims}
    \item[] Question: Do the main claims made in the abstract and introduction accurately reflect the paper's contributions and scope?
\item[] Answer: \answerYes{}
\item[] Justification: The abstract and introduction state the main contributions and scope, and the claims are supported by the method and experimental results.

\item {\bf Limitations}
    \item[] Question: Does the paper discuss the limitations of the work performed by the authors?
\item[] Answer: \answerYes{}
\item[] Justification: We included it in the paper.

\item {\bf Theory assumptions and proofs}
    \item[] Question: For each theoretical result, does the paper provide the full set of assumptions and a complete (and correct) proof?
\item[] Answer: \answerYes{}
\item[] Justification: The assumptions and derivation for the first-order analysis of null-space gradient projection are provided in the appendix.

    \item {\bf Experimental result reproducibility}
    \item[] Question: Does the paper fully disclose all the information needed to reproduce the main experimental results of the paper to the extent that it affects the main claims and/or conclusions of the paper (regardless of whether the code and data are provided or not)?
\item[] Answer: \answerYes{}
\item[] Justification: The paper provides all the necessary details to reproduce the results.

\item {\bf Open access to data and code}
    \item[] Question: Does the paper provide open access to the data and code, with sufficient instructions to faithfully reproduce the main experimental results, as described in supplemental material?
\item[] Answer: \answerYes{} 
\item[] Justification: The source code is publicly available at: \url{https://github.com/LAMDA-CL/NeurIPS2026-Drape}. Detailed algorithmic descriptions are provided in the appendix.

\item {\bf Experimental setting/details}
    \item[] Question: Does the paper specify all the training and test details (e.g., data splits, hyperparameters, how they were chosen, type of optimizer) necessary to understand the results?
    \item[] Answer: \answerYes{} 
    \item[] Justification: The experiment details are available in the paper.

\item {\bf Experiment statistical significance}
    \item[] Question: Does the paper report error bars suitably and correctly defined or other appropriate information about the statistical significance of the experiments?
\item[] Answer: \answerNo{}
\item[] Justification: The paper does not report error bars, but follows standard MCIT evaluation protocols on fixed benchmarks and shows consistent gains across multiple baselines. Repeated full MCIT runs are computationally expensive.

\item {\bf Experiments compute resources}
    \item[] Question: For each experiment, does the paper provide sufficient information on the computer resources (type of compute workers, memory, time of execution) needed to reproduce the experiments?
\item[] Answer: \answerYes{}
\item[] Justification: The paper reports the computational resources used for the experiments.
    
\item {\bf Code of ethics}
    \item[] Question: Does the research conducted in the paper conform, in every respect, with the NeurIPS Code of Ethics \url{https://neurips.cc/public/EthicsGuidelines}?
\item[] Answer: \answerYes{}
\item[] Justification: The research conforms to the NeurIPS Code of Ethics.

\item {\bf Broader impacts}
    \item[] Question: Does the paper discuss both potential positive societal impacts and negative societal impacts of the work performed?
\item[] Answer: \answerYes{}
\item[] Justification: The paper discusses potential positive impacts and general risks of multimodal continual adaptation in appendix.
    
\item {\bf Safeguards}
    \item[] Question: Does the paper describe safeguards that have been put in place for responsible release of data or models that have a high risk for misuse (e.g., pre-trained language models, image generators, or scraped datasets)?
\item[] Answer: \answerNA{}
\item[] Justification: No high-risk data or models are released.

\item {\bf Licenses for existing assets}
    \item[] Question: Are the creators or original owners of assets (e.g., code, data, models), used in the paper, properly credited and are the license and terms of use explicitly mentioned and properly respected?
\item[] Answer: \answerYes{}
\item[] Justification: All the datasets used are publicly available, discussed, and properly cited in Section~\ref{sec:exp}.

\item {\bf New assets}
    \item[] Question: Are new assets introduced in the paper well documented and is the documentation provided alongside the assets?
\item[] Answer: \answerNA{}
\item[] Justification: The paper does not release new datasets, models, or benchmarks.

\item {\bf Crowdsourcing and research with human subjects}
    \item[] Question: For crowdsourcing experiments and research with human subjects, does the paper include the full text of instructions given to participants and screenshots, if applicable, as well as details about compensation (if any)? 
\item[] Answer: \answerNA{}
\item[] Justification: The paper does not involve crowdsourcing or new human-subject studies.

\item {\bf Institutional review board (IRB) approvals or equivalent for research with human subjects}
    \item[] Question: Does the paper describe potential risks incurred by study participants, whether such risks were disclosed to the subjects, and whether Institutional Review Board (IRB) approvals (or an equivalent approval/review based on the requirements of your country or institution) were obtained?
\item[] Answer: \answerNA{}
\item[] Justification: The paper does not involve human-subject research.

\item {\bf Declaration of LLM usage}
    \item[] Question: Does the paper describe the usage of LLMs if it is an important, original, or non-standard component of the core methods in this research? Note that if the LLM is used only for writing, editing, or formatting purposes and does \emph{not} impact the core methodology, scientific rigor, or originality of the research, declaration is not required.
\item[] Answer: \answerNA{}
\item[] Justification: The core method development in this research does not involve LLMs

\end{enumerate}

\end{document}